\pdfoutput=1

\documentclass[11pt,table]{article}

\usepackage[final]{misc/acl}

\usepackage{times}
\usepackage{latexsym}
\usepackage[T1]{fontenc}
\usepackage[utf8]{inputenc}
\usepackage{multirow}
\usepackage{enumitem}
\usepackage{microtype}
\usepackage{amssymb}
\usepackage{amsmath}
\usepackage{inconsolata}
\usepackage{booktabs}
\usepackage{tabularx} 
\usepackage{graphicx}
\usepackage{cleveref}
\usepackage{listings}

\lstdefinestyle{promptstyle}{
 basicstyle=\ttfamily\footnotesize,
 backgroundcolor=\color{gray!10},
 breaklines=true,
 frame=single,
 columns=fullflexible,
 breakindent=0pt
}

\definecolor{TodoColor}{rgb}{1,0.7,0.6}

\makeatletter\def\Hy@Warning#1{}\makeatother
\let\svthefootnote\thefootnote
\newcommand\blankfootnote[1]{%
  \let\thefootnote\relax\footnotetext{#1}%
  \let\thefootnote\svthefootnote%
}

\crefname{example}{example}{examples}
\Crefname{example}{Example}{Examples}
\DeclareCaptionType{example}[Example][List of examples]

\usepackage{tikz}

\usepackage{todonotes}
\usepackage{marginnote}

\title{Multilingual Performance Biases of Large Language Models in Education}

\title{Are Large Language Models for Education Reliable for All Languages?}

\author{
\bf Vansh Gupta$\thanks{Equal Contribution}^\dagger$ 
\qquad
\bf Sankalan Pal Chowdhury$^{*\dagger}$
\\
\bf Vilém Zouhar$^\dagger$
\qquad
\bf Donya Rooein$^\ddagger$
\qquad
\bf Mrinmaya Sachan$^\dagger$ 
\\[0.5em]
\tt\small \{\href{mailto:guptav@ethz.ch}{\color{black} guptav},\href{mailto:spalchowd@ethz.ch}{\color{black} spalchowd},\href{mailto:vzouhar@ethz.ch}{\color{black} vzouhar},\href{mailto:msachan@ethz.ch}{\color{black} msachan}\}@ethz.ch
\quad
\href{mailto:donya.rooein@unibocconi.it}{\color{black} donya.rooein@unibocconi.it}
\\
$^\dagger$ETH Zurich \quad $^\ddagger$Bocconi University
}

\begin{document}
\maketitle
\begin{abstract}
Large language models (LLMs) are increasingly being adopted in educational settings.
These applications expand beyond English, though current LLMs remain primarily English-centric.
In this work, we ascertain if their use in education settings in non-English languages is warranted. We evaluated the performance of popular LLMs on four educational tasks: identifying student misconceptions, providing targeted feedback, interactive tutoring, and grading translations in eight languages (Mandarin, Hindi, Arabic, German, Farsi, Telugu, Ukrainian, Czech) in addition to English.
We find that the performance on these tasks somewhat corresponds to the amount of language represented in training data, with lower-resource languages having poorer task performance.
However, at least some models are able to more or less maintain their levels of performance across all languages.
Thus, we recommend that practitioners first verify that the LLM works well in the target language for their educational task before deployment.
\end{abstract}

\footnotetext[0]{
We release the collected dataset and code at \href{https://github.com/eth-lre/multilingual-educational-llm-bias}{github.com/}
\href{https://github.com/eth-lre/multilingual-educational-llm-bias}{eth-lre/multilingual-educational-llm-bias}.
The dataset comprises 313,500 automatically evaluated model outputs across seven languages, four tasks, and six models.
}

\vspace{-2mm}
\section{Introduction}


Education is a multilingual, multicultural endeavour. 
AI-based technologies have recently shown the potential to improve students' learning experiences, and educational systems worldwide are increasingly adopting these tools \cite{Gligorea2023AdaptiveLUA}.
From personalized instruction and targeted feedback to appropriate content generation and interactive tutoring, these tools offer solutions to key educational challenges \cite{Leon2024LeveragingGAA, rooein-etal-2024-beyond,Mosher2024ThePPA}.
Large language models such as GPT, Gemini, and Llama \citep{openai2023gpt4, geminiteam2024geminifamilyhighlycapable,202307.2142} have become particularly influential, with early evidence suggesting their ability to support teachers or scaffold student learning \cite{Kasneci2023ChatGPTFGA, Alqahtani2023TheERA}.

Although most of these LLMs are trained on multilingual corpora \citep{openai2019language,nvidia_transformer,peng2023rwkv,gu2023mamba}, they are still overwhelmingly English-centric \citep[][\Cref{tab:language details}]{Argoub_2022, ruder-etal-2022-square}.
Inadequate adaptation to local languages in an educational setting risks diminishing their utility and exacerbating existing inequalities by privileging dominant languages and cultures.
The question of multilingualism arises in every domain where LLMs are applied \cite{lai-etal-2023-chatgpt, ahuja-etal-2023-mega, ahuja-etal-2024-megaverse}. However, it is especially important in the field of education, which has seen wide use of LLMs despite the high stakes \citep{alhafni2024llmseducationnovelperspectives,raheja2023coedittexteditingtaskspecific,naismith-etal-2023-automated}.
Without rigorous evaluation tailored to educational tasks across languages, deploying LLMs in classrooms may introduce new forms of harm, including misinformation, misalignment with curricula, or culturally inappropriate content \citep{Almasoud2025TowardIEA}.

In this work, we present an empirical investigation of the capabilities of frontier LLMs on educational tasks across several languages.
We identify four education-related tasks (identifying student misconceptions, providing targeted feedback, interactive tutoring, and translation grading) with well-defined language-agnostic metrics.
We then evaluate several frontier LLMs (Claude, Gemini, GPT4o, Llama, and Mistral) on these tasks in eight languages (Mandarin, Hindi, Arabic, German, Farsi, Telugu, Ukrainian, and Czech) in addition to English.

Our results show that though performance in English still dominates, other languages are not too far behind, at least for GPT4o and Gemini-2.0-flash, which emerge as the best models. We also find that using prompts in the language of the task is rarely helpful compared to English prompts.

\begin{table*}[t]
 \setlength{\tabcolsep}{10pt}
 \small
 \centering
 \begin{tabular}{rrrrrrr}
 \toprule
 &
 \bf Language family &
 \bf Script &
 \bf Wikipedia &
 \bf CommonCrawl & 
 \bf Speakers \\
 \midrule
 English & Germanic & Latin & 6973K & 42.8\% & 1500M \\
 Mandarin & Sino-tibetan & Hanzi\footnotemark & 1480K & 5.8\% & 1184M \\
 Hindi & Indo-Iranian & Brahmic & 165K & 0.20\% & 609M \\
 Arabic & Afro-Asiatic & Abjad & 1259K & 0.68\% & 411M \\
 German & Germanic & Latin & 3021K & 5.5\% & 411M \\
 Farsi & Indo-Iranian & Abjad & 1034K & 0.74\% & 134M \\
 Telugu & Dravidian & Brahmic & 111K & 0.02\% & 96M \\
 Ukrainian & Slavic & Cyrillic & 1371K & 0.62\% & 39M \\
 Czech & Slavic & Latin & 566M & 0.10\% & 12M \\
 \bottomrule
 \end{tabular}

\vspace{-2mm}
\caption{
Language information, number of speakers (\href{https://www.ethnologue.com/}{Ethnologue 2025}), and global representations of tested languages in NLP (Wikipedia Articles and proportion in CommonCrawl in March 2025).
}
\label{tab:language details}
\end{table*}

\section{Methods}
\label{sec:methods}

We select our set of tasks based on 3 desiderata:
\begin{itemize}[left=0mm]
    \item \textbf{Relevant to Education:} We focus on tasks that LLMs would encounter specifically in the role of tutors, teachers, or teaching assistants. We do not cover tasks like question answering or solving math questions, which, while possibly being relevant to education, are more general tasks that are primarily studied in other contexts.
    \item \textbf{Have a Language Component:} We avoid tasks whose formulation uses purely notation, for example, solving a math equation. If the equation is provided in mathematical notation, the task would remain unchanged between different languages, making the question of multilingual performance moot.
    \item \textbf{Language Invariant Evaluation:} Finally, we need evaluation metrics that remain comparable across languages to compare performance across different languages efficiently. This means we cannot rely on language-dependent metrics like  BLEU or COMET \citep{papineni-etal-2002-bleu,rei-etal-2020-comet}.
\end{itemize}

\noindent
Based on these, we selected following four tasks:

\paragraph{Task 1: Misconception identification.}
An important aspect of teaching is fixing student misconceptions, which first requires identifying the student misconception \citep{liu2023novicelearnerexperttutor}. We build this task on the \href{https://eedi.com/us}{EEDI} Math Questions Dataset, which contains thousands of multiple-choice questions with four answer choices. For many of the wrong choices, we have expert-annotated misconceptions that could lead to a student picking the said choice. We leverage these to build our task. The LLM is given a multiple-choice question, the student's (incorrect) answer, and four possible misconceptions.
The candidate misconceptions include the true misconception identified by experts and three distractors chosen at random from the other misconceptions present in the dataset. The LLM must pick the correct misconception from these four options (see \Cref{example:misconception} for an example). 
We evaluate the LLM performance by reporting accuracy in predicting the student misconception. Since the model must pick one of four options, a random baseline has an accuracy of 25\%.

\paragraph{Task 2: Feedback selection.}
A key step towards fixing students' misconceptions is generating feedback to alleviate them.
The \href{https://eedi.com/us}{EEDI} dataset discussed above also includes feedback for all the choices we use for this part. The LLM is again given a multiple-choice question, the student's answer, and this time, a set of four possible feedbacks, out of which the LLM must select the feedback corresponding to the student's answer. Note that while there are $4$ possible feedbacks, one corresponds to the correct answer. This one is easily identifiable as it reinforces the student's answer, while the feedbacks corresponding to wrong answers all try to make the student realize their mistake. As an example, see Option C in both parts of \ref{example:feedback}, which are the only options in their respective questions that do not start with a negative tone. Therefore, if the selected answer is also the correct answer to the problem, the LLM might be able to pick the correct feedback using some shallow semantics, which we want to avoid. Therefore, we ensure that the selected answer is always incorrect. The random baseline has an accuracy of 25\%, or 33\% if choosing among responses to the wrong answer.
%

\paragraph{Task 3: Tutoring.}\label{par:tut}
For more complex misconceptions, a single-turn feedback often does not suffice, and fixing the misconception requires a multi-turn conversation between the student and the teacher, also known as tutoring. \cite{doi:10.3102/0013189x013006004, ee9d2bdb-d01b-3241-a32a-984f3da551e0}
This involves a teacher LLM trying to help the student identify and fix an error in their solution. We evaluate the tutoring ability of the LLM by having it tutor a weaker LLM, which acts as the student. 
Both the teacher and the student are given the question, but only the teacher LLM can access the correct answer.
The student LLM is instructed to stick to the wrong solution unless it sees strong justification to shift. The teacher and the student take turns to send messages, with the teacher's goal being to get the student model to the correct answer, without revealing the answer themselves. The teacher LLM is considered to get a \textit{success} if the student LLM states the answer. If the teacher reveals the answer before the student has gotten to it, it is counted as \textit{telling}. An \textit{adjusted success} occurs when there is a success but no telling. The task is finally evaluated by Tutoring score \citep{10.1145/3657604.3662041}, which is the harmonic mean between success rate and adjusted success rate.

This task differs from the other tasks on this list in at least two significant ways. First, it is a multi-turn conversation task, so there is no scope for guessing the answer. Secondly, the final evaluation depends on the performance of the student LLM, so the multilingual capabilities of the student LLM also restrict the applicability of this task. These factors make this task both slower to run and more complex for the LLMs.

\paragraph{Task 4: Translation grading.}
A common field of education that has seen an increase in the use of LLMs is Language learning \cite{klimova2024exploring, zhu2024embracingaieducationunderstanding}. A representative task from this field is to assign a grade to a translation provided by a student. While we lack proper datasets across languages with translations and their appropriate grades, we can approximate this task by the fact that \textit{the machine translation of a sentence should receive a higher grade than the exact translation with one word replaced by a random word.}
\footnotetext{Alternately referred to as Kanji, Hanja or Hantu}
We use English sentences from Duolingo's English$\rightarrow$Spanish SLAM dataset \citep{DVN/8SWHNO_2018}, which are machine translated to other languages. We chose this dataset because it is meant to be used for translation, so it should contain fewer hard-to-translate sentences.
We filter out simple sentences that do not end with a full stop or have fewer than five words.
For each translated sentence, we then create a corresponding \textit{perturbed translation} by replacing one of the words in the sentence with a different word selected at random from the other sentences in the dataset, disrupting both the fluency and adequacy of the translation. 
The LLM judges both the original and perturbed versions on a scale from 1 (completely incorrect) to 5 (perfect), with the expectation that it should assign a strictly lower score to the perturbed version.
A model assigning all scores at random would therefore score around 40\%.

\begin{figure}[ht]
    \centering
    \includegraphics[width=\columnwidth]{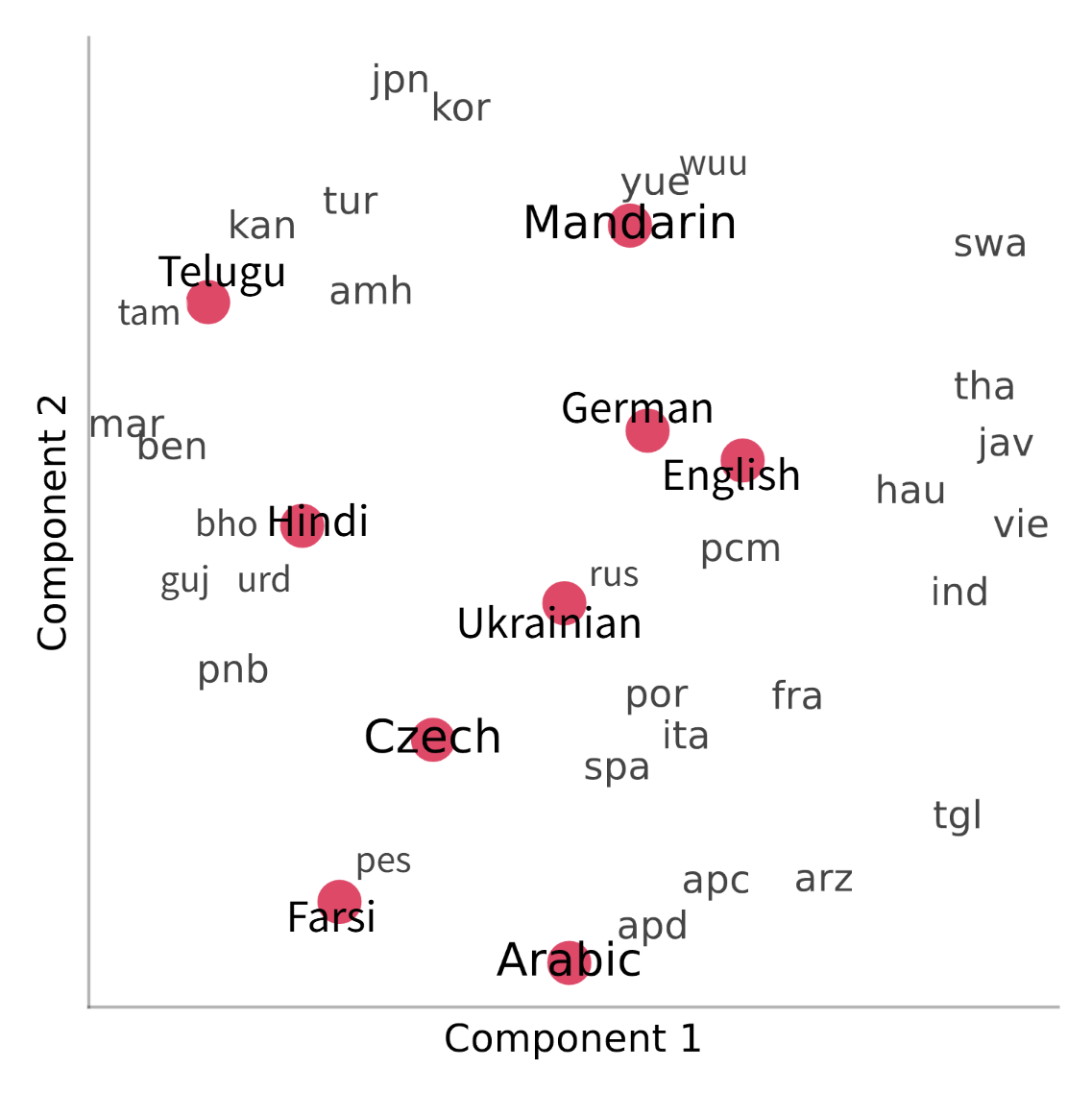}
    \caption{Multidimensional Scaling projection of languages based on syntax features from URIEL/lang2vec. Languages used in our experiments are highlighted and shown with full names, others are in ISO 639/set 2.}
    \label{fig:langsim}
\end{figure}
\begin{table*}[!h]
\centering \small
\begin{tabular}{lcccc}
\toprule
\bf Language
& \textbf{Questions}
& \textbf{Misconception}
& \textbf{Feedback}
& \textbf{Translation} \\
\midrule
{Mandarin} & 0.593&0.607&0.666&0.781\\
{Hindi} & 0.455& 0.546 & 0.593 & 0.831 \\
{Arabic} & 0.596& 0.659 & 0.605 & 0.793 \\
{German} & 0.623 &0.642&0.697&0.792\\
{Farsi} & 0.574& 0.644 & 0.708 & 0.840 \\
{Telugu}& 0.518& 0.569 & 0.578 & 0.639 \\
{Ukrainian} & 0.607& 0.642 & 0.682 & 0.821 \\
{Czech} & 0.611& 0.626 & 0.663 & 0.805 \\
\bottomrule
\end{tabular}
\caption{Average COMET$^\text{DA,XL}_\text{23}$ scores for different languages for different components of the tasks. The \textbf{Questions} are used for both Misconception and Feedback tasks. The Tutoring task is not translated.}
\label{tab:comet}
\end{table*}
\paragraph{Language selection.}
We choose eight languages for experiments: Mandarin, Hindi, Arabic, German, Farsi, Telugu, Ukranian, Czech in addition to English for comparison.
This language selection reflects diverse linguistic properties, varying levels of representation in training data, and different language families \citep{wikimedia_list_wikipedias}, see \Cref{tab:language details}.
Hindi (Indo-Aryan) and Telugu (Dravidian) represent major languages from the Indian subcontinent that use the Brahmic script and are under-represented in both CommonCrawl and Wikipedia. German and Mandarin, on the other hand, are examples of languages well represented in both CommonCrawl and Wikipedia
Farsi and Arabic offer insights into LLM performance on a right-to-left Abjad script, whereas Ukrainian and Czech allow us to study generalisation in medium resource morphologically rich languages, using the Cyrillic and Latin scripts, respectively.

\begin{figure*}[ht!]
\begin{minipage}{\linewidth}
 \centering
 \small
 \centering
 \begin{tabularx}{\linewidth}{p{2.9cm}X}
   \toprule
   \textbf{Input} & \textbf{Options} \\
   \midrule
   \textbf{Question:} Which number is the greatest? \newline
   \textbf{Student Answer:}\newline 5.0001 \newline\textbf{Right Answer}:5.2&
   \textbf{A:} Believes the mean is total frequency divided by something, \newline
   \textbf{B (correct):} Thinks the more digits a number has the greater it is, regardless of place value, \newline
   \textbf{C:} Believes parallel lines have gradients that multiply to give -1, \newline
   \textbf{D:} When multiplying by a multiple of 10, gives an answer 10 times bigger than it should be\\
   & \\[-0.1em]
   \textbf{Question:} What is the lowest common multiple of \(8\) and \(4\)? \newline
   \textbf{Student answer:} \(4\)\newline \textbf{Right Answer:} 8&
   \textbf{A:} Subtracts instead of adds when answering worded problems, \newline
   \textbf{B (correct):} Confuses factors and multiples, \newline
   \textbf{C:} Rounds up instead of down, \newline
   \textbf{D:} Adds instead of multiplying when expanding bracket \\
   \bottomrule
 \end{tabularx}

 \vspace{-2mm}
 \captionof{example}{Two examples of the misconception identification task (English).}
 \label{example:misconception}

\vspace{4mm}

 \centering
 \small \centering
 \begin{tabularx}{\linewidth}{p{2.5cm}X}
   \toprule
   \textbf{Input} & \textbf{Options} \\
   \midrule
   \textbf{Question:} 6 pencils cost £1.50. How much do 3 pencils cost? \newline
   \textbf{Student answer:} 25p
   & \textbf{A:} I think you have made an arithmetic error when halving £1.50. Use short division to divide by two,\newline
   \textbf{B:} I think you have used the incorrect notation for money. Consider how the monetary values in the question are written, \newline
   \textbf{C (correct answer):} If 6 pencils cost £1.50, then 3 pencils cost half of £1.50, which is £0.75 or 75p., \newline
   \textbf{D (student answer):} I think you have found the cost for one pencil. The question asks for the cost of 3 pencils.  \\
   & \\[-0.1em]
   \textbf{Question:} A film starts at 8.50pm. The film lasts 2 hours and 52 minutes. What time does the film finish? \newline
   \textbf{Student answer:} \newline 11.02pm &
   \textbf{A (student answer):} This isn't quite right. Remember that there are 60 minutes in an hour, not 100 :), \newline
   \textbf{B:} I think you've confused your method a little. Noticing that 2 hours and 52 minutes is just 8 minutes less than 3 hours is super, just make sure you add and subtract in the correct directions though :), \newline
   \textbf{C:} Almost there! Take care to notice how many hours and minutes you're adding here. Is your answer 2 hours and 52 minutes later than 8.50pm?, \newline
   \textbf{D (correct answer):} Adding 2 hours to 8.50pm gives 10.50pm. Adding 10 minutes on takes us to 11.00pm, and adding the remaining 42 minutes gives 11.42pm.\\
   \bottomrule
 \end{tabularx}

 \vspace{-2mm}
 \captionof{example}{Two examples of the feedback selection task (English).}
 \label{example:feedback}

 \vspace{4mm}

 \centering
 \small \centering
 \begin{tabularx}{\linewidth}{p{5.8cm} X p{2.9cm}}
   \toprule
   \textbf{Math Problem} & \textbf{Student's (Incorrect) Solution} & \textbf{Correct Solution}\\
   \midrule
   Sam sells bread. He has a target of selling 120 crates of bread in a week. One week he was closed on Monday and Friday. Over the weekend he sold 20 crates. On Tuesday he sold 15 crates, on Wednesday 12 crates, and Thursday 18 crates. By how many crates was Sam off from his target for the week? 
   & Sam had 5 days to sell bread because he was closed on Monday and Friday. He sold a total of 20 + 15 + 12 + 18 = 65 crates of bread from Tuesday to Thursday. Adding the 20 crates he sold over the weekend, Sam sold a total of 65 + 20 = 85 crates of bread in a week. Sam was off from his target by 120 - 85 = 35 crates of bread.  
   & During the whole week Sam sold 15 + 12 + 18 + 20 = 65 crates. Sam was off his target by 120 - 65 = 55 crates. \\
   && \\[-0.1em]
   Sophia is thinking of taking a road trip in her car, and would like to know how far she can drive on a single tank of gas. She has traveled 100 miles since last filling her tank, and she needed to put in 4 gallons of gas to fill it up again. The owner's manual for her car says that her tank holds 12 gallons of gas. How many miles can Sophia drive on a single tank of gas?  
   & Sophia used 4 out of the 12 gallons of gas in her tank, so there are 12-4 = 8 gallons of gas left in the tank. If Sophia can drive 100 miles on 4 gallons of gas, then she can drive 100/4 = 25 miles per gallon. Therefore, with 8 gallons of gas left in the tank, Sophia can drive 25 x 8 = 200 miles on a single tank of gas.  
   & To find miles per gallon, divide 100 miles / 4 gallons = 25 miles per gallon. To find how far Olivia can go on a single tank, multiply 25 miles per gallon × 12 gallons = 300 miles.  \\
   \bottomrule
 \end{tabularx}%

 \vspace{-2mm}
 \captionof{example}{Two examples of the tutoring task.}
 \label{example:tutoring}
 
\vspace{4mm}

 \centering
 \small \centering
 \begin{tabularx}{\linewidth}{llllX}
   \toprule
   \textbf{English Source} & \textbf{Original Translation} & \textbf{Perturbed Translation} & \textbf{Language} \\
   \midrule
   \multirow{8}{*}{It is a kind of tomato.}  
   & \includegraphics[height=1em]{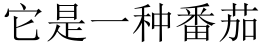}
   & \includegraphics[height=1em]{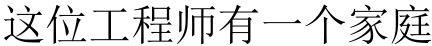}
   & Mandarin\\
   & \includegraphics[height=1em]{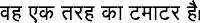}
   & \includegraphics[height=1em]{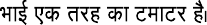}
   & Hindi \\
   & \includegraphics[height=1em]{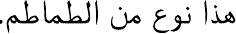}
   & \includegraphics[height=1em]{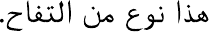}
   & Arabic \\
   & Es ist eine Art Tomate
   & Katze ist eine Art Tomate
   & German \\
   & \includegraphics[height=1em]{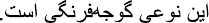}
   & \includegraphics[height=1em]{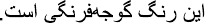}
   & Farsi \\
   & \includegraphics[height=1em]{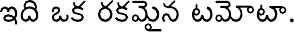}
   & \includegraphics[height=1em]{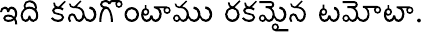}
   & Telugu \\
   & \includegraphics[height=0.9em]{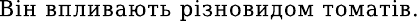}
   & \includegraphics[height=0.9em]{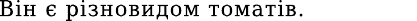} 
   \hspace{-1.5cm}
   & Ukrainian \\
   & Je to druh rajčete.  
   & matka to druh rajčete.  
   & Czech \\
   \bottomrule
 \end{tabularx}
 
 \vspace{-2mm}
 \captionof{example}{A single example of the translation grading task for non-English languages.}
 \label{example:translation}

\end{minipage}
\end{figure*}

To assess the typological diversity of our selected languages, we used the {URIEL typological database} \cite{littell-etal-2017-uriel} with \href{https://pypi.org/project/lang2vec/}{\texttt{lang2vec}}, which provides dense vector representations of languages based on a range of typological, phylogenetic, and geographical features.
As recommended by the package, we extracted \texttt{syntax} features with k-NN predictions for the missing values for a set of 40 languages, constructed as the union of our core experimental languages and the most widely spoken languages worldwide according to Ethnologue \cite{ethnologue2025}.
We projected each language feature vector into two dimensions using Multidimensional Scaling, producing a 2D language similarity plot.
This allows us to visualise (see \Cref{fig:langsim}) the relative syntactic diversity of our selected languages and confirm that they span a broad typological space.
The visualisation demonstrates that our language selection (highlighted) is well distributed across the typological landscape.

\paragraph{Translation.}
We obtain our tasks in all the above-mentioned languages by machine translation. Following the GPT4 Technical Report \cite[][Figure 5]{openai2023gpt4}, we use Azure Translate to translate all our examples to the target languages. However, this introduces an additional noise source for tasks performed in languages other than English. In fact, after reviewing some of the translations manually, it does look like the translations, though decent, are not as easy to follow as their English counterparts. This finding is further corroborated by COMET$^\text{DA,XL}_\text{23}$ \cite{rei-etal-2023-scaling} scores of the translations (see \Cref{tab:comet}).
This means that any differences we observe between English and other-language performance cannot be conclusively attributed to the LLM being tested. However, we can still compare the performance of different LLMs across the same language, as the same translation was used for all LLMs. Further, if at least one LLM performs well in a task on a given language, we can be reasonably certain that the translation for that task-language pair was also good enough. 


\paragraph{Models and prompts.}
We evaluate six state-of-the-art LLMs praised for their multilingual capabilities:
GPT-4o \cite{openai2023gpt4},
Gemini 2.0 Flash \cite{geminiteam2024geminifamilyhighlycapable},
Claude 3.7 Sonnet \cite{anthropic_claude_3_5},
Llama 3.1 405B \cite{grattafiori2024llama3herdmodels},
Mistral Large 2407 \cite{mistral_large_2407,jiang2023mistral}, and Command-A \cite{cohere2025command}. We leave all sampling parameters to their defaults. 
For prompts, we use a simple chain of thought prompting method, where the model is first asked to explain why it would pick a certain answer, and then asked to choose it in a separate prompt.
Based on literature \cite{mondshine-etal-2024-hesum, huang2023languagescreatedequalllms}, it is unclear whether or not it is beneficial to translate the prompt itself to the target language or keep it in English, so we try both options.%
\footnote{
A weaker model roleplays the student model used in the tutoring task to be consistent with the original work.
We only use the original prompts because it does not work well with non-English prompts.
}\textsuperscript{,}%
\footnote{We machine-translate the prompts and manually verify (with L1/L2 language knowledge) the translation adequacy.}

For each task, we use 1000 examples for reporting our results, sampled at random from the dataset, except for 200 examples in the tutoring task, which is multi-turn.


\begin{table*}[t!]
\fontsize{7.5}{9}\selectfont \centering

{
\setlength{\tabcolsep}{3.4pt}
\begin{tabular}{ lrrrrrr@{\hspace{2mm}}crrrrrr }
\toprule
& \multicolumn{ 6 }{c}{\bf English prompt} && \multicolumn{ 6 }{c}{\bf Translated prompt} \\
\textbf{Language} & \bf GPT4o & \bf LLama & \bf Claude & \bf Gemini & \bf Mistral & \bf Cmd-A & \bf  & \bf GPT4o & \bf LLama & \bf Claude & \bf Gemini & \bf Mistral & \bf Cmd-A\\
\midrule
English & \cellcolor{purple!48} 97.6\% & \cellcolor{purple!47} 96.2\% & \cellcolor{purple!46} 95.1\% & \cellcolor{purple!46} 94.0\% & \cellcolor{purple!46} 95.0\% & \cellcolor{purple!47} 95.3\% &  & \cellcolor{purple!48} 97.6\% & \cellcolor{purple!47} 96.2\% & \cellcolor{purple!46} 95.1\% & \cellcolor{purple!46} 94.0\% & \cellcolor{purple!46} 95.0\% & \cellcolor{purple!47} 95.3\%\\
Mandarin & \cellcolor{purple!47} $\cdot$ 95.8\% & \cellcolor{purple!47} 95.2\% & \cellcolor{purple!45} $\cdot$ 92.5\% & \cellcolor{purple!44} $\cdot$ 92.1\% & \cellcolor{purple!45} $\cdot$ 92.8\% & \cellcolor{purple!45} $\cdot$ 92.9\% &  & \cellcolor{purple!48} 96.5\% & \cellcolor{purple!46} 95.0\% & \cellcolor{purple!44} $\cdot$ 91.7\% & \cellcolor{purple!46} 93.8\% & \cellcolor{purple!45} $\cdot$ 92.9\% & \cellcolor{purple!46} 94.1\%\\
Hindi & \cellcolor{purple!46} $\cdot$ 94.5\% & \cellcolor{purple!45} $\cdot$ 93.2\% & \cellcolor{purple!44} $\cdot$ 91.9\% & \cellcolor{purple!43} $\cdot$ 89.8\% & \cellcolor{purple!44} $\cdot$ 91.8\% & \cellcolor{purple!45} $\cdot$ 93.2\% &  & \cellcolor{purple!47} $\cdot$ 95.5\% & \cellcolor{purple!45} $\cdot$ 93.6\% & \cellcolor{purple!43} $\cdot$ 89.6\% & \cellcolor{purple!43} $\cdot$ 90.4\% & \cellcolor{purple!43} $\cdot$ 90.6\% & \cellcolor{purple!44} $\cdot$ 91.4\%\\
Arabic & \cellcolor{purple!47} $\cdot$ 95.9\% & \cellcolor{purple!45} $\cdot$ 93.0\% & \cellcolor{purple!44} $\cdot$ 92.0\% & \cellcolor{purple!40} $\star$ 86.0\% & \cellcolor{purple!45} $\cdot$ 92.6\% & \cellcolor{purple!45} $\cdot$ 93.4\% &  & \cellcolor{purple!47} $\cdot$ 95.9\% & \cellcolor{purple!45} $\cdot$ 93.0\% & \cellcolor{purple!45} $\cdot$ 92.8\% & \cellcolor{purple!44} $\cdot$ 90.9\% & \cellcolor{purple!44} $\cdot$ 92.0\% & \cellcolor{purple!46} 94.0\%\\
German & \cellcolor{purple!47} $\cdot$ 96.0\% & \cellcolor{purple!47} 96.2\% & \cellcolor{purple!46} 94.6\% & \cellcolor{purple!39} $\star$ 84.6\% & \cellcolor{purple!46} 95.1\% & \cellcolor{purple!47} 95.2\% &  & \cellcolor{purple!47} $\cdot$ 95.9\% & \cellcolor{purple!48} 96.6\% & \cellcolor{purple!46} 94.0\% & \cellcolor{purple!31} $\bigstar$74.0\% & \cellcolor{purple!46} 94.9\% & \cellcolor{purple!47} 95.2\%\\
Farsi & \cellcolor{purple!46} $\cdot$ 94.8\% & \cellcolor{purple!45} $\cdot$ 93.3\% & \cellcolor{purple!45} $\cdot$ 93.0\% & \cellcolor{purple!41} $\cdot$ 87.5\% & \cellcolor{purple!45} $\cdot$ 92.7\% & \cellcolor{purple!45} $\cdot$ 93.1\% &  & \cellcolor{purple!46} $\cdot$ 95.1\% & \cellcolor{purple!46} $\cdot$ 94.4\% & \cellcolor{purple!27} $\bigstar$68.0\% & \cellcolor{purple!42} $\cdot$ 88.3\% & \cellcolor{purple!26} $\bigstar$66.9\% & \cellcolor{purple!45} $\cdot$ 93.6\%\\
Telugu & \cellcolor{purple!47} $\cdot$ 95.2\% & \cellcolor{purple!44} $\cdot$ 92.2\% & \cellcolor{purple!43} $\cdot$ 89.9\% & \cellcolor{purple!41} $\cdot$ 86.9\% & \cellcolor{purple!43} $\cdot$ 89.7\% & \cellcolor{purple!40} $\star$ 85.5\% &  & \cellcolor{purple!46} $\cdot$ 94.2\% & \cellcolor{purple!43} $\cdot$ 90.8\% & \cellcolor{purple!28} $\bigstar$68.6\% & \cellcolor{purple!38} $\star$ 83.6\% & \cellcolor{purple!4} $\bigstar$35.5\% & \cellcolor{purple!34} $\bigstar$77.9\%\\
Ukranian & \cellcolor{purple!47} $\cdot$ 95.7\% & \cellcolor{purple!46} 94.9\% & \cellcolor{purple!45} $\cdot$ 92.9\% & \cellcolor{purple!45} 93.3\% & \cellcolor{purple!46} 94.4\% & \cellcolor{purple!46} 94.9\% &  & \cellcolor{purple!47} $\cdot$ 95.6\% & \cellcolor{purple!46} $\cdot$ 94.3\% & \cellcolor{purple!19} $\bigstar$56.6\% & \cellcolor{purple!43} $\cdot$ 90.4\% & \cellcolor{purple!46} 94.2\% & \cellcolor{purple!46} 93.9\%\\
Czech & \cellcolor{purple!48} 96.9\% & \cellcolor{purple!46} 95.1\% & \cellcolor{purple!46} 94.5\% & \cellcolor{purple!44} 92.3\% & \cellcolor{purple!46} 94.5\% & \cellcolor{purple!46} 94.1\% &  & \cellcolor{purple!48} 96.6\% & \cellcolor{purple!47} 95.8\% & \cellcolor{purple!29} $\bigstar$70.2\% & \cellcolor{purple!37} $\star$ 81.6\% & \cellcolor{purple!8} $\bigstar$41.0\% & \cellcolor{purple!46} 94.5\%\\
\bottomrule
\end{tabular}
}

\vspace{-2.5mm}
\caption{Results (accuracy) for the \textbf{misconception identification} task.
We mark results significantly lower (at least 10\%=$\bigstar$, at least 5\%=$\star$, otherwise $\cdot$) than English with a one-sided 95\% confidence t-test.
}
\label{tab:result_misconception}
\medskip

{
\setlength{\tabcolsep}{3.4pt}
\begin{tabular}{ lrrrrrr@{\hspace{2mm}}crrrrrr }
\toprule
& \multicolumn{ 6 }{c}{\bf English prompt} && \multicolumn{ 6 }{c}{\bf Translated prompt} \\
\textbf{Language} & \bf GPT4o & \bf LLama & \bf Claude & \bf Gemini & \bf Mistral & \bf Cmd-A & \bf  & \bf GPT4o & \bf LLama & \bf Claude & \bf Gemini & \bf Mistral & \bf Cmd-A\\
\midrule
English & \cellcolor{cyan!44} 53.4\% & \cellcolor{cyan!31} 38.2\% & \cellcolor{cyan!13} 17.0\% & \cellcolor{cyan!42} 51.1\% & \cellcolor{cyan!40} 48.5\% & \cellcolor{cyan!32} 39.7\% &  & \cellcolor{cyan!44} 53.4\% & \cellcolor{cyan!31} 38.2\% & \cellcolor{cyan!13} 17.0\% & \cellcolor{cyan!42} 51.1\% & \cellcolor{cyan!40} 48.5\% & \cellcolor{cyan!32} 39.7\%\\
Mandarin & \cellcolor{cyan!41} $\cdot$ 49.6\% & \cellcolor{cyan!24} $\star$ 29.7\% & \cellcolor{cyan!9} $\cdot$ 12.3\% & \cellcolor{cyan!35} $\cdot$ 43.0\% & \cellcolor{cyan!33} $\cdot$ 40.1\% & \cellcolor{cyan!26} $\cdot$ 31.8\% &  & \cellcolor{cyan!34} $\star$ 41.1\% & \cellcolor{cyan!15} $\bigstar$19.2\% & \cellcolor{cyan!3} $\star$ 5.8\% & \cellcolor{cyan!24} $\bigstar$30.3\% & \cellcolor{cyan!24} $\bigstar$30.3\% & \cellcolor{cyan!22} $\star$ 27.8\%\\
Hindi & \cellcolor{cyan!40} $\cdot$ 48.7\% & \cellcolor{cyan!29} 35.6\% & \cellcolor{cyan!9} $\cdot$ 13.0\% & \cellcolor{cyan!36} $\cdot$ 43.6\% & \cellcolor{cyan!33} $\cdot$ 40.5\% & \cellcolor{cyan!26} $\cdot$ 31.6\% &  & \cellcolor{cyan!26} $\bigstar$32.1\% & \cellcolor{cyan!10} $\bigstar$13.4\% & \cellcolor{cyan!4} $\star$ 6.2\% & \cellcolor{cyan!36} $\cdot$ 44.3\% & \cellcolor{cyan!14} $\bigstar$18.6\% & \cellcolor{cyan!14} $\bigstar$18.8\%\\
Arabic & \cellcolor{cyan!41} $\cdot$ 49.6\% & \cellcolor{cyan!23} $\star$ 28.7\% & \cellcolor{cyan!10} $\cdot$ 13.9\% & \cellcolor{cyan!37} $\cdot$ 45.3\% & \cellcolor{cyan!32} $\star$ 38.8\% & \cellcolor{cyan!27} $\cdot$ 33.3\% &  & \cellcolor{cyan!40} $\cdot$ 48.8\% & \cellcolor{cyan!8} $\bigstar$10.7\% & \cellcolor{cyan!12} 16.3\% & \cellcolor{cyan!40} 48.1\% & \cellcolor{cyan!22} $\bigstar$27.8\% & \cellcolor{cyan!23} $\star$ 28.9\%\\
German & \cellcolor{cyan!44} 52.5\% & \cellcolor{cyan!26} $\cdot$ 32.1\% & \cellcolor{cyan!11} 15.0\% & \cellcolor{cyan!38} $\cdot$ 46.4\% & \cellcolor{cyan!35} $\cdot$ 42.4\% & \cellcolor{cyan!27} $\cdot$ 32.8\% &  & \cellcolor{cyan!42} 50.6\% & \cellcolor{cyan!25} $\cdot$ 30.8\% & \cellcolor{cyan!12} 15.6\% & \cellcolor{cyan!37} $\cdot$ 44.4\% & \cellcolor{cyan!32} $\star$ 39.4\% & \cellcolor{cyan!31} 37.6\%\\
Farsi & \cellcolor{cyan!42} 50.2\% & \cellcolor{cyan!22} $\star$ 27.9\% & \cellcolor{cyan!8} $\cdot$ 11.3\% & \cellcolor{cyan!37} $\cdot$ 44.9\% & \cellcolor{cyan!34} $\cdot$ 41.3\% & \cellcolor{cyan!25} $\star$ 30.9\% &  & \cellcolor{cyan!38} $\cdot$ 45.9\% & \cellcolor{cyan!26} $\cdot$ 31.6\% & \cellcolor{cyan!12} 16.3\% & \cellcolor{cyan!36} $\cdot$ 44.0\% & \cellcolor{cyan!27} $\bigstar$33.5\% & \cellcolor{cyan!29} $\cdot$ 35.5\%\\
Telugu & \cellcolor{cyan!37} $\cdot$ 45.2\% & \cellcolor{cyan!22} $\star$ 27.6\% & \cellcolor{cyan!7} $\cdot$ 10.4\% & \cellcolor{cyan!36} $\cdot$ 43.4\% & \cellcolor{cyan!28} $\bigstar$34.0\% & \cellcolor{cyan!21} $\star$ 26.3\% &  & \cellcolor{cyan!10} $\bigstar$13.9\% & \cellcolor{cyan!9} $\bigstar$12.7\% & \cellcolor{cyan!4} $\star$ 6.1\% & \cellcolor{cyan!31} $\star$ 37.7\% & \cellcolor{cyan!12} $\bigstar$15.5\% & \cellcolor{cyan!6} $\bigstar$9.5\%\\
Ukranian & \cellcolor{cyan!42} 50.3\% & \cellcolor{cyan!27} $\cdot$ 33.2\% & \cellcolor{cyan!9} $\cdot$ 13.0\% & \cellcolor{cyan!37} $\cdot$ 44.8\% & \cellcolor{cyan!34} $\cdot$ 41.3\% & \cellcolor{cyan!26} $\cdot$ 32.2\% &  & \cellcolor{cyan!29} $\bigstar$35.9\% & \cellcolor{cyan!15} $\bigstar$19.6\% & \cellcolor{cyan!5} $\star$ 8.1\% & \cellcolor{cyan!44} 52.8\% & \cellcolor{cyan!25} $\bigstar$31.0\% & \cellcolor{cyan!22} $\star$ 27.2\%\\
Czech & \cellcolor{cyan!41} 49.9\% & \cellcolor{cyan!31} 37.8\% & \cellcolor{cyan!10} $\cdot$ 14.1\% & \cellcolor{cyan!38} $\cdot$ 46.5\% & \cellcolor{cyan!34} $\cdot$ 41.6\% & \cellcolor{cyan!25} $\star$ 30.7\% &  & \cellcolor{cyan!35} $\star$ 42.7\% & \cellcolor{cyan!21} $\star$ 26.1\% & \cellcolor{cyan!15} 19.2\% & \cellcolor{cyan!38} $\cdot$ 46.6\% & \cellcolor{cyan!29} $\star$ 35.5\% & \cellcolor{cyan!29} $\cdot$ 35.6\%\\
\bottomrule
\end{tabular}
}

\vspace{-2.5mm}
\caption{Results (accuracy) for the \textbf{feedback selection} task.
We mark results significantly lower (at least 10\%=$\bigstar$, at least 5\%=$\star$, otherwise $\cdot$) than English with a one-sided 95\% confidence t-test.
}
\label{tab:result_feedback}

\medskip

{
\setlength{\tabcolsep}{1.7pt}
\begin{tabular}{ lrrrrrr@{\hspace{1mm}}crrrrrr }
\toprule
& \multicolumn{ 6 }{c}{\bf Harmonic mean} && \multicolumn{ 6 }{c}{\bf Success/1-Telling} \\
 & \bf GPT4o & \bf LLama & \bf Claude & \bf Gemini & \bf Mistral & \bf Cmd-A & \bf  & \bf GPT4o & \bf LLama & \bf Claude & \bf Gemini & \bf Mistral & \bf Cmd-A\\
\midrule
English & \cellcolor{orange!47} 94.7\% & \cellcolor{orange!48} 97.0\% & \cellcolor{orange!11} 22.1\% & \cellcolor{orange!46} 93.0\% & \cellcolor{orange!41} 82.0\% & \cellcolor{orange!48} 95.5\% &  & \cellcolor{orange!47} 96.0/2.5\% & \cellcolor{orange!48} 97.5/1.0\% & \cellcolor{orange!11} 96.5/84.0\% & \cellcolor{orange!46} 93.5/1.0\% & \cellcolor{orange!41} 82.0/0.0\% & \cellcolor{orange!48} 96.0/1.0\%\\
Mandarin & \cellcolor{orange!45} 89.8\% & \cellcolor{orange!44} 89.0\% & \cellcolor{orange!13} 26.4\% & \cellcolor{orange!40} 79.7\% & \cellcolor{orange!40} 79.7\% & \cellcolor{orange!44} 88.2\% &  & \cellcolor{orange!45} 94.0/8.0\% & \cellcolor{orange!44} 90.5/3.0\% & \cellcolor{orange!13} 90.0/74.5\% & \cellcolor{orange!40} 80.5/1.5\% & \cellcolor{orange!40} 80.0/0.5\% & \cellcolor{orange!44} 93.0/9.0\%\\
Hindi & \cellcolor{orange!45} 90.5\% & \cellcolor{orange!46} 92.7\% & \cellcolor{orange!12} 24.2\% & \cellcolor{orange!36} $\star$ 72.2\% & \cellcolor{orange!37} 73.5\% & \cellcolor{orange!44} $\cdot$ 88.4\% &  & \cellcolor{orange!45} 95.0/8.5\% & \cellcolor{orange!46} 93.0/0.5\% & \cellcolor{orange!12} 89.5/75.5\% & \cellcolor{orange!36} 77.5/10.0\% & \cellcolor{orange!37} 73.5/0.0\% & \cellcolor{orange!44} 91.0/5.0\%\\
Arabic & \cellcolor{orange!46} 91.4\% & \cellcolor{orange!45} 89.7\% & \cellcolor{orange!12} 24.3\% & \cellcolor{orange!42} $\cdot$ 84.2\% & \cellcolor{orange!38} 75.2\% & \cellcolor{orange!44} 87.4\% &  & \cellcolor{orange!46} 94.5/5.9\% & \cellcolor{orange!45} 90.0/0.5\% & \cellcolor{orange!12} 91.0/77.0\% & \cellcolor{orange!42} 86.0/3.5\% & \cellcolor{orange!38} 75.5/0.5\% & \cellcolor{orange!44} 93.0/10.5\%\\
German & \cellcolor{orange!45} 90.7\% & \cellcolor{orange!46} 91.2\% & \cellcolor{orange!12} 23.4\% & \cellcolor{orange!42} 84.2\% & \cellcolor{orange!39} 77.2\% & \cellcolor{orange!43} $\cdot$ 86.3\% &  & \cellcolor{orange!45} 92.5/3.5\% & \cellcolor{orange!46} 92.0/1.5\% & \cellcolor{orange!12} 88.0/74.5\% & \cellcolor{orange!42} 85.0/1.5\% & \cellcolor{orange!39} 77.5/0.5\% & \cellcolor{orange!43} 90.5/8.0\%\\
Farsi & \cellcolor{orange!43} $\cdot$ 85.6\% & \cellcolor{orange!41} $\star$ 81.3\% & \cellcolor{orange!14} 28.7\% & \cellcolor{orange!39} 77.2\% & \cellcolor{orange!33} $\cdot$ 65.8\% & \cellcolor{orange!39} $\cdot$ 77.8\% &  & \cellcolor{orange!43} 89.0/6.5\% & \cellcolor{orange!41} 87.5/11.5\% & \cellcolor{orange!14} 91.5/74.5\% & \cellcolor{orange!39} 78.0/1.5\% & \cellcolor{orange!33} 69.5/7.0\% & \cellcolor{orange!39} 91.0/23.0\%\\
Telugu & \cellcolor{orange!25} $\bigstar$50.1\% & \cellcolor{orange!20} $\bigstar$39.5\% & \cellcolor{orange!14} 27.7\% & \cellcolor{orange!29} $\cdot$ 58.9\% & \cellcolor{orange!1} $\bigstar$2.9\% & \cellcolor{orange!20} $\bigstar$40.7\% &  & \cellcolor{orange!25} 77.5/40.5\% & \cellcolor{orange!20} 77.5/51.0\% & \cellcolor{orange!14} 85.5/69.0\% & \cellcolor{orange!29} 61.0/4.0\% & \cellcolor{orange!1} 59.0/57.5\% & \cellcolor{orange!20} 63.5/33.5\%\\
Ukranian & \cellcolor{orange!46} 91.2\% & \cellcolor{orange!46} 91.5\% & \cellcolor{orange!12} 23.5\% & \cellcolor{orange!41} $\cdot$ 81.2\% & \cellcolor{orange!36} 71.5\% & \cellcolor{orange!45} 90.9\% &  & \cellcolor{orange!46} 93.0/3.5\% & \cellcolor{orange!46} 92.0/1.0\% & \cellcolor{orange!12} 91.5/78.0\% & \cellcolor{orange!41} 84.0/5.5\% & \cellcolor{orange!36} 71.5/0.0\% & \cellcolor{orange!45} 93.5/5.0\%\\
Czech & \cellcolor{orange!22} $\bigstar$43.8\% & \cellcolor{orange!22} $\bigstar$44.1\% & \cellcolor{orange!9} 17.2\% & \cellcolor{orange!35} 70.2\% & \cellcolor{orange!1} $\bigstar$2.9\% & \cellcolor{orange!11} $\bigstar$21.5\% &  & \cellcolor{orange!22} 65.5/32.5\% & \cellcolor{orange!22} 73.5/42.0\% & \cellcolor{orange!9} 90.0/80.5\% & \cellcolor{orange!35} 71.5/2.5\% & \cellcolor{orange!1} 52.5/51.0\% & \cellcolor{orange!11} 77.0/64.5\%\\
\bottomrule
\end{tabular}
}

\vspace{-2.5mm}
\caption{Results (harmonic mean, success, and telling) for the \textbf{tutoring} task.
We mark results significantly lower (at least 10\%=$\bigstar$, at least 5\%=$\star$, otherwise $\cdot$) than English with a one-sided 95\% confidence t-test when occurring in both success and telling.
Telling is flipped such that higher is better.
}
\label{tab:result_tutoring}

\medskip

{
\setlength{\tabcolsep}{5.2pt}
\begin{tabular}{ lrrrrrr@{\hspace{2mm}}crrrrrr }
\toprule
& \multicolumn{ 6 }{c}{\bf English prompt} && \multicolumn{ 6 }{c}{\bf Translated prompt} \\
\textbf{Language} & \bf GPT4o & \bf LLama & \bf Claude & \bf Gemini & \bf Mistral & \bf Cmd-A & \bf  & \bf GPT4o & \bf LLama & \bf Claude & \bf Gemini & \bf Mistral & \bf Cmd-A\\
\midrule
Mandarin & \cellcolor{green!50} 100.0\% & \cellcolor{green!50} 99.3\% & \cellcolor{green!49} 98.9\% & \cellcolor{green!50} 99.9\% & \cellcolor{green!50} 99.5\% & \cellcolor{green!50} 99.9\% &  & \cellcolor{green!50} 99.9\% & \cellcolor{green!50} 99.4\% & \cellcolor{green!6} 24.8\% & \cellcolor{green!50} 99.6\% & \cellcolor{green!50} 99.4\% & \cellcolor{green!50} 99.9\%\\
Hindi & \cellcolor{green!45} 91.5\% & \cellcolor{green!35} 74.1\% & \cellcolor{green!45} 92.1\% & \cellcolor{green!37} 77.6\% & \cellcolor{green!40} 82.4\% & \cellcolor{green!37} 77.9\% &  & \cellcolor{green!46} 93.8\% & \cellcolor{green!43} 88.5\% & \cellcolor{green!24} 56.5\% & \cellcolor{green!42} 86.5\% & \cellcolor{green!43} 87.6\% & \cellcolor{green!39} 81.3\%\\
Arabic & \cellcolor{green!49} 98.6\% & \cellcolor{green!49} 97.9\% & \cellcolor{green!50} 99.2\% & \cellcolor{green!49} 98.8\% & \cellcolor{green!49} 97.5\% & \cellcolor{green!49} 99.0\% &  & \cellcolor{green!49} 98.8\% & \cellcolor{green!49} 98.3\% & \cellcolor{green!31} 67.2\% & \cellcolor{green!49} 98.6\% & \cellcolor{green!49} 97.8\% & \cellcolor{green!49} 97.9\%\\
German & \cellcolor{green!49} 98.2\% & \cellcolor{green!49} 97.9\% & \cellcolor{green!49} 97.9\% & \cellcolor{green!49} 98.2\% & \cellcolor{green!49} 98.2\% & \cellcolor{green!49} 98.2\% &  & \cellcolor{green!49} 98.5\% & \cellcolor{green!49} 98.3\% & \cellcolor{green!9} 29.9\% & \cellcolor{green!49} 98.0\% & \cellcolor{green!49} 98.3\% & \cellcolor{green!49} 97.8\%\\
Farsi & \cellcolor{green!47} 95.3\% & \cellcolor{green!46} 93.5\% & \cellcolor{green!48} 96.0\% & \cellcolor{green!48} 96.4\% & \cellcolor{green!45} 92.3\% & \cellcolor{green!48} 96.6\% &  & \cellcolor{green!48} 96.8\% & \cellcolor{green!48} 96.0\% & \cellcolor{green!31} 67.0\% & \cellcolor{green!48} 96.4\% & \cellcolor{green!47} 94.1\% & \cellcolor{green!48} 96.2\%\\
Telugu & \cellcolor{green!37} 77.2\% & \cellcolor{green!11} 33.7\% & \cellcolor{green!39} 81.0\% & \cellcolor{green!22} 51.9\% & \cellcolor{green!20} 48.7\% & \cellcolor{green!6} 25.2\% &  & \cellcolor{green!40} 82.8\% & \cellcolor{green!19} 46.8\% & \cellcolor{green!15} 40.7\% & \cellcolor{green!39} 82.1\% & \cellcolor{green!31} 67.1\% & \cellcolor{green!0} 15.6\%\\
Ukranian & \cellcolor{green!49} 98.0\% & \cellcolor{green!48} 97.3\% & \cellcolor{green!48} 96.9\% & \cellcolor{green!48} 96.5\% & \cellcolor{green!48} 97.3\% & \cellcolor{green!49} 98.3\% &  & \cellcolor{green!49} 98.1\% & \cellcolor{green!49} 97.9\% & \cellcolor{green!41} 85.3\% & \cellcolor{green!49} 97.7\% & \cellcolor{green!49} 98.4\% & \cellcolor{green!49} 98.2\%\\
Czech & \cellcolor{green!49} 98.7\% & \cellcolor{green!49} 98.3\% & \cellcolor{green!49} 98.9\% & \cellcolor{green!49} 98.3\% & \cellcolor{green!49} 97.5\% & \cellcolor{green!49} 98.8\% &  & \cellcolor{green!50} 99.3\% & \cellcolor{green!49} 98.8\% & \cellcolor{green!39} 80.8\% & \cellcolor{green!49} 98.7\% & \cellcolor{green!50} 99.5\% & \cellcolor{green!50} 99.2\%\\
\bottomrule
\end{tabular}
}

\vspace{-2.5mm}
\caption{Results (accuracy) for the \textbf{translation grading} task.}
\label{tab:result_translation}

\vspace{-3.5mm}
\end{table*}
\begin{table*}[t]
\centering \small
\begin{tabular}{lcccc}
\toprule
\bf Language
& \textbf{Questions}
& \textbf{Misconception}
& \textbf{Feedback}
& \textbf{Translation} \\
\midrule
{Mandarin} & 0.593&0.607&0.666&0.781\\
{Hindi} & 0.455& 0.546 & 0.593 & 0.831 \\
{Arabic} & 0.596& 0.659 & 0.605 & 0.793 \\
{German} & 0.623 &0.642&0.697&0.792\\
{Farsi} & 0.574& 0.644 & 0.708 & 0.840 \\
{Telugu}& 0.518& 0.569 & 0.578 & 0.639 \\
{Ukrainian} & 0.607& 0.642 & 0.682 & 0.821 \\
{Czech} & 0.611& 0.626 & 0.663 & 0.805 \\
\bottomrule
\end{tabular}
\caption{Average COMET$^\text{DA,XL}_\text{23}$ scores for different languages for different components of the tasks. The \textbf{Questions} are used for both Misconception and Feedback tasks. The Tutoring task is not translated.}
\label{tab:comet}
\end{table*}
\section{Results}

In this section, we describe the results of five popular large language models on the four tasks described in \Cref{sec:methods}.
The main results are shown in 
\Cref{tab:result_misconception,tab:result_feedback,tab:result_translation,tab:result_tutoring}.

\paragraph{English is easiest for LLMs.}
The gap between English and other languages is large in general.
On average\footnote{Averaging here is done to give a general idea, but we must note that the scores are not equivalent. We use Accuracy for tasks 1 and 2 but Tutoring Score for Task 3} across all tasks (excluding translation) and models, English has 70.9\%, in contrast to 63.1\% (Hindi), 55.3\% (Czech), 67.8\% (Ukrainian), 49.7\% (Telugu), 66.2\% (Farsi), 66.8\% (German), 64.6\% (Mandarin) and 67.4\% (Arabic).
This in itself does not make it clear if the loss is due to the LLMs being weak or the translation quality being poor. The poor performance on Telugu is largely driven by Command-A and Mistral. The former is unsurprising as Telugu is the only language in our list that is not officially supported by it \cite{cohere2025command}. On the other hand, Mistral lists only \href{https://mistral.ai/news/mistral-large-2407}{12 supported languages} of which we test only Hindi, Arabic, German and Chinese. Telugu also has the lowest representation in CommonCrawl and Wikipedia, so the result is expected.
Manual analysis of the low tutoring performance for Czech reveals that the interactions switch between various language formality styles, to the point that it becomes distracting.
Additionally, the language used in Czech classrooms is particular and likely not represented on the internet.

\paragraph{Model performance and consistency.}
Mistral is the most inconsistent across non-English languages (average deviation\footnote{\label{fn:fn1}We calculate the standard deviation across the six languages for each task and then calculate the mean.}=0.186).
For example, it completely fails the tutoring task for both Czech and Telugu, despite performing reasonably on other languages in the same task. Command-A is not much better (average deviation\textsuperscript{\ref{fn:fn1}}=0.161). 
On the other hand, Gemini is the most consistent (average deviation\textsuperscript{\ref{fn:fn1}}=0.078) and also has the second-best performance (average score 75.0\%).
GPT4o, is the best performing model (average 78.6\%) while Claude performs the worst (average 49.3\%) mostly due to Feedback and Tutoring tasks.

\paragraph{Task difficulty.}
The worst performance is observed in the Feedback task despite the similarity to the Misconception identification task.
While Claude is still the standout worst performer with a worse-than-random performance, all models struggle.
Further analysis in \Cref{tab:default_correct} shows that all models tended to default the feedback corresponding to the correct answer, with the models' chain of thoughts being ``regardless of the student's mistake, this is the feedback that gives the student the most information about the correct answer.''
Most models perform well in the Translation evaluation task, with the accuracy being even higher than human annotators, who were presented with attention checks with similar perturbations \citep{kocmi-etal-2024-error,zouhar2025ai}. They also do well in the Misconceptions task, with most percentage scores (at least in the English prompt setting) being in the 90s. 
The tutoring task seems to have the most inconsistent performance across models and languages. In general, all models struggle in Czech and Telugu, while Claude struggles in all languages. Avoiding telling seems to be the more challenging part of the problem for all the models, although success rates are not very consistent either.



\paragraph{English and translated prompts.}

Excluding for the Tutoring task (which did not use native prompts), using English prompts yields better performance than using translated prompts (averages 72.7\% and 67.2\%). The exceptions to these are GPT, Llama, Gemini, and Mistral in the translation task though in most cases, the difference is not very large. Note that some of the poor performance could be attributed to the prompts being translated and checked for correctness rather than being written in the target language directly, which could introduce some translationese. Regardless, we believe it is best to keep prompts in English. As a further note for English-speaking developers designing multilingual applications, keeping prompts in English ensures that the chains-of-thought remain English, making it easier to run sanity checks.

\section{Related Work}

LLMs, trained on vast multilingual texts, have dominated tasks such as text generation, translation, and dialogue \cite{brown2020language}, making them promising tools in Intelligent Tutoring Systems \citep[ITS;][]{corbett1997849, 10.1145/3657604.3662041}.
Prior work explores their use in educational contexts, such as dynamic student interactions \cite{schmucker2023ruffleriley}, simulating expert and novice behavior \cite{liu2023novicelearnerexperttutor}, and math word problem reasoning \cite{opedal2023world}.

Beyond mathematical context, LLMs have also been explored for other forms of learning. \citet{cui-sachan-2023-adaptive} investigate LLMs in adaptive and personalized exercise generation for language learners, while \cite{wang-etal-2023-strategize} examines how conversational tutoring strategies can aid student understanding.
Additionally, LLMs have been used to assess grammatical correctness and translation accuracy \cite{kocmi-federmann-2023-large,omelianchuk-etal-2024-pillars,freitag-etal-2024-llms}, facilitate automated essay scoring \cite{PACK2024100234}, and provide corrective feedback in second language writing \cite{han-etal-2024-llm}.
\begin{table*}[!h]
\centering
\small
\begin{tabular}{llrrrrr}
\toprule
\bf Model  & \bf API & \bf Total & \bf Miconception & \bf Feedback & \bf Tutoring & \bf Translation \\
\midrule
Mistral & Mistral API & \$530 & \$170 & \$170 & \$120 & \$70 \\
Claude  & Anthropic   & \$600 & \$190 & \$190 & \$135 & \$85 \\
Command  & Cohere   & \$520 & \$165 & \$165 & \$120 & \$70 \\
Llama   & Together.ai & \$600 &  \$190 & \$190 & \$135 & \$80 \\
GPT4o   & Open AI& \$80  & \$25 & \$25 & \$18 & \$12  \\
Gemini  & Google Genai & \$30  & \$10 & \$10 & \$6 & \$4 \\
\bottomrule
\end{tabular}
\caption{Approximate costs for the experiments. Does not include taxes or currency conversion charges. The total is about \$2360 with approximately an additional \$500 spent on preliminary experiments.}
\label{tab:cost}
\end{table*}
While LLMs excel in English, their abilities in other languages often vary, reflecting an over-representation of high-resource languages in pre-training corpora.
For example, \citet{koto2023large} introduces IndoMMLU, which reveals significant performance disparities between Indonesian and English contexts.
Similarly, \citet{holtermann2024evaluatingelementarymultilingualcapabilities} examines LLMs across 137 languages and attributes discrepancies in performance to tokenisation strategies. \citet{li2024quantifyingmultilingualperformancelarge,armengol-estape-etal-2022-multilingual} further find a strong correlation between pre-training data proportions and performance, reaffirming the gap between high- and low-resource languages.
For Catalan, 
\citet{armengol-estape-etal-2022-multilingual} find that while GPT-3 performed well in generative tasks, its comprehension capabilities were limited by the language's moderate representation.

Recent research has increasingly explored the application of LLMs in multilingual educational contexts, though challenges persist in balancing performance across languages. Systematic reviews of AI-based language learning tools highlight the prevalence of NLP and machine learning techniques for error correction, feedback provision, and assessment in non-English contexts, though they note persistent gaps in dialogic competence and teacher preparedness \cite{alhusaiyan2025systematic}. Studies evaluating LLMs’ cross-lingual capabilities reveal performance disparities, with models demonstrating stronger skill tagging accuracy for English-centric curricula compared to underrepresented languages like Irish or Marathi \cite{https://doi.org/10.1111/bjet.13465}. Bibliometric analyses indicate growing research interest in AI for foreign language education, particularly in vocabulary acquisition and writing support, though most studies still focus on high-resource European and Asian languages \cite{dougan2024artificial}. These works collectively underscore both the transformative potential and current limitations of LLMs in achieving equitable multilingual educational support.

To address multilingual education more directly, projects like Kaleidoscope \cite{salazar2025kaleidoscopeinlanguageexamsmassively} and Aya \cite{ustun-etal-2024-aya} by Cohere For AI aim to support culturally diverse languages, while SEA-HELM \cite{susanto2025seahelmsoutheastasianholistic} and ECLeKTic \cite{goldman2025eclekticnovelchallengeset} emphasise culturally grounded evaluations in Southeast Asian and cross-lingual contexts, respectively. These efforts highlight the need for multilingual benchmarks that move beyond English-centric evaluations.

Prior pedagogical studies tend to assess single LLMs in monolingual settings.
We fill this gap by benchmarking LLMs in multiple tasks. 
Specifically, we conduct zero-shot experiments across multiple models and languages to better analyze their real-world applicability.


\section{Conclusion}

We analyse the performance of six well-known state-of-the-art LLMs across six languages other than English on four educational tasks.
We find that while performance in English continues to be better than in other languages, the drop to other models is not always large. In particular, we find that GPT4o and Gemini 2.0 perform consistently well across all languages, with a few exceptions. We also note that English prompts work as well, if not better, than prompts written in the target language, when solving multilingual tasks. This opens up opportunities for porting applications developed for English into different languages. However, we note that certain models perform poorly in some tasks and languages, so \textbf{we recommend} first verifying that a model works well in a particular language on a specific educational task before deployment. However, to answer the question posed by the title, we believe that \textit{atleast some} language models \textbf{are} reliable across languages.

%


\section*{Limitations}

The shown experiments could naturally be better extended to more languages.
The selected languages reflect a balance between author familiarity, which is necessary for meaningful qualitative analysis, and linguistic diversity, as evidenced by their spread in URIEL feature space.
Similarly, we only covered six LLMs.
In both cases, the cost of experiments (see \Cref{tab:cost}) becomes prohibitively expensive, which motivated the data release in this paper to enable further research.

Additionally, translation quality remains a concern, as previously discussed. A more thorough evaluation would involve human translations for every task, similar to the MMLU multilingual benchmark \cite{xuan2025mmluproxmultilingualbenchmarkadvanced}, but doing so for all our tasks would be resource-intensive.

Finally, the set of tasks is not a complete representation of problems in the education space, primarily because most of the more complex tasks lack well-defined language-agnostic metrics.
\section*{Acknowledgements}

Sankalan Pal Chowdhury is partially funded by the ETH-EPFL JDPLS Program. Donya Rooein is supported by the European Research Council (ERC) under the European Union’s Horizon 2020 research and innovation program (grant agreement No. 949944, INTEGRATOR).

\bibliography{misc/anthology.min,misc/bibliography}

\clearpage

\appendix

\onecolumn

\begin{table*}[ht]
\vspace{-2mm}

\fontsize{7.5}{9}\selectfont \centering
\setlength{\tabcolsep}{5pt}

\begin{tabular}{ lrrrrrr@{\hspace{2mm}}crrrrrr }
\toprule
& \multicolumn{ 6 }{c}{\bf English prompt} && \multicolumn{ 6 }{c}{\bf Translated prompt} \\
\textbf{Language} & \bf GPT4o & \bf LLama & \bf Claude & \bf Gemini & \bf Mistral & \bf Cmd-A & \bf  & \bf GPT4o & \bf LLama & \bf Claude & \bf Gemini & \bf Mistral & \bf Cmd-A\\
\midrule
English & \cellcolor{purple!0} 0.0\% & \cellcolor{purple!7} 0.7\% & \cellcolor{purple!1} 0.1\% & \cellcolor{purple!21} 2.1\% & \cellcolor{purple!0} 0.0\% & \cellcolor{purple!9} 0.9\% &  & \cellcolor{purple!0} 0.0\% & \cellcolor{purple!7} 0.7\% & \cellcolor{purple!1} 0.1\% & \cellcolor{purple!21} 2.1\% & \cellcolor{purple!0} 0.0\% & \cellcolor{purple!9} 0.9\%\\
Mandarin & \cellcolor{purple!5} 0.5\% & \cellcolor{purple!16} 1.6\% & \cellcolor{purple!0} 0.0\% & \cellcolor{purple!32} 3.2\% & \cellcolor{purple!0} 0.0\% & \cellcolor{purple!2} 0.2\% &  & \cellcolor{purple!5} 0.5\% & \cellcolor{purple!16} 1.6\% & \cellcolor{purple!0} 0.0\% & \cellcolor{purple!32} 3.2\% & \cellcolor{purple!0} 0.0\% & \cellcolor{purple!2} 0.2\%\\
Hindi & \cellcolor{purple!0} 0.0\% & \cellcolor{purple!16} 1.6\% & \cellcolor{purple!4} 0.4\% & \cellcolor{purple!23} 2.3\% & \cellcolor{purple!0} 0.0\% & \cellcolor{purple!4} 0.4\% &  & \cellcolor{purple!0} 0.0\% & \cellcolor{purple!9} 0.9\% & \cellcolor{purple!2} 0.2\% & \cellcolor{purple!25} 2.5\% & \cellcolor{purple!0} 0.0\% & \cellcolor{purple!1} 0.1\%\\
Arabic & \cellcolor{purple!1} 0.1\% & \cellcolor{purple!17} 1.7\% & \cellcolor{purple!2} 0.2\% & \cellcolor{purple!21} 2.1\% & \cellcolor{purple!0} 0.0\% & \cellcolor{purple!2} 0.2\% &  & \cellcolor{purple!0} 0.0\% & \cellcolor{purple!11} 1.1\% & \cellcolor{purple!3} 0.3\% & \cellcolor{purple!22} 2.2\% & \cellcolor{purple!0} 0.0\% & \cellcolor{purple!1} 0.1\%\\
German & \cellcolor{purple!5} 0.5\% & \cellcolor{purple!16} 1.6\% & \cellcolor{purple!3} 0.3\% & \cellcolor{purple!23} 2.3\% & \cellcolor{purple!0} 0.0\% & \cellcolor{purple!2} 0.2\% &  & \cellcolor{purple!5} 0.5\% & \cellcolor{purple!16} 1.6\% & \cellcolor{purple!3} 0.3\% & \cellcolor{purple!23} 2.3\% & \cellcolor{purple!0} 0.0\% & \cellcolor{purple!2} 0.2\%\\
Farsi & \cellcolor{purple!0} 0.0\% & \cellcolor{purple!18} 1.8\% & \cellcolor{purple!2} 0.2\% & \cellcolor{purple!20} 2.0\% & \cellcolor{purple!0} 0.0\% & \cellcolor{purple!3} 0.3\% &  & \cellcolor{purple!0} 0.0\% & \cellcolor{purple!16} 1.6\% & \cellcolor{purple!2} 0.2\% & \cellcolor{purple!29} 2.9\% & \cellcolor{purple!0} 0.0\% & \cellcolor{purple!1} 0.1\%\\
Telugu & \cellcolor{purple!0} 0.0\% & \cellcolor{purple!1} 0.1\% & \cellcolor{purple!0} 0.0\% & \cellcolor{purple!22} 2.2\% & \cellcolor{purple!0} 0.0\% & \cellcolor{purple!4} 0.4\% &  & \cellcolor{purple!0} 0.0\% & \cellcolor{purple!3} 0.3\% & \cellcolor{purple!1} 0.1\% & \cellcolor{purple!17} 1.7\% & \cellcolor{purple!0} 0.0\% & \cellcolor{purple!0} 0.0\%\\
Ukranian & \cellcolor{purple!1} 0.1\% & \cellcolor{purple!16} 1.6\% & \cellcolor{purple!1} 0.1\% & \cellcolor{purple!22} 2.2\% & \cellcolor{purple!0} 0.0\% & \cellcolor{purple!3} 0.3\% &  & \cellcolor{purple!0} 0.0\% & \cellcolor{purple!18} 1.8\% & \cellcolor{purple!4} 0.4\% & \cellcolor{purple!16} 1.6\% & \cellcolor{purple!0} 0.0\% & \cellcolor{purple!1} 0.1\%\\
Czech & \cellcolor{purple!1} 0.1\% & \cellcolor{purple!16} 1.6\% & \cellcolor{purple!1} 0.1\% & \cellcolor{purple!19} 1.9\% & \cellcolor{purple!0} 0.0\% & \cellcolor{purple!7} 0.7\% &  & \cellcolor{purple!0} 0.0\% & \cellcolor{purple!14} 1.4\% & \cellcolor{purple!0} 0.0\% & \cellcolor{purple!7} 0.7\% & \cellcolor{purple!0} 0.0\% & \cellcolor{purple!5} 0.5\%\\
\bottomrule
\end{tabular}

\vspace{-2mm}
\caption{Response error rate for the \textbf{misconception identification} task.}
\medskip

\begin{tabular}{ lrrrrrr@{\hspace{2mm}}crrrrrr }
\toprule
& \multicolumn{ 6 }{c}{\bf English prompt} && \multicolumn{ 6 }{c}{\bf Translated prompt} \\
\textbf{Language} & \bf GPT4o & \bf LLama & \bf Claude & \bf Gemini & \bf Mistral & \bf Cmd-A & \bf  & \bf GPT4o & \bf LLama & \bf Claude & \bf Gemini & \bf Mistral & \bf Cmd-A\\
\midrule
English & \cellcolor{cyan!0} 0.0\% & \cellcolor{cyan!8} 0.3\% & \cellcolor{cyan!0} 0.0\% & \cellcolor{cyan!32} 1.3\% & \cellcolor{cyan!0} 0.0\% & \cellcolor{cyan!0} 0.0\% &  & \cellcolor{cyan!0} 0.0\% & \cellcolor{cyan!8} 0.3\% & \cellcolor{cyan!0} 0.0\% & \cellcolor{cyan!32} 1.3\% & \cellcolor{cyan!0} 0.0\% & \cellcolor{cyan!0} 0.0\%\\
Mandarin & \cellcolor{cyan!0} 0.0\% & \cellcolor{cyan!2} 0.1\% & \cellcolor{cyan!0} 0.0\% & \cellcolor{cyan!38} 1.5\% & \cellcolor{cyan!0} 0.0\% & \cellcolor{cyan!5} 0.2\% &  & \cellcolor{cyan!0} 0.0\% & \cellcolor{cyan!0} 0.0\% & \cellcolor{cyan!2} 0.1\% & \cellcolor{cyan!40} 1.6\% & \cellcolor{cyan!0} 0.0\% & \cellcolor{cyan!2} 0.1\%\\
Hindi & \cellcolor{cyan!0} 0.0\% & \cellcolor{cyan!0} 0.0\% & \cellcolor{cyan!0} 0.0\% & \cellcolor{cyan!27} 1.1\% & \cellcolor{cyan!0} 0.0\% & \cellcolor{cyan!2} 0.1\% &  & \cellcolor{cyan!0} 0.0\% & \cellcolor{cyan!0} 0.0\% & \cellcolor{cyan!0} 0.0\% & \cellcolor{cyan!25} 1.0\% & \cellcolor{cyan!0} 0.0\% & \cellcolor{cyan!5} 0.2\%\\
Arabic & \cellcolor{cyan!0} 0.0\% & \cellcolor{cyan!0} 0.0\% & \cellcolor{cyan!0} 0.0\% & \cellcolor{cyan!38} 1.5\% & \cellcolor{cyan!0} 0.0\% & \cellcolor{cyan!2} 0.1\% &  & \cellcolor{cyan!0} 0.0\% & \cellcolor{cyan!0} 0.0\% & \cellcolor{cyan!0} 0.0\% & \cellcolor{cyan!52} 2.1\% & \cellcolor{cyan!0} 0.0\% & \cellcolor{cyan!5} 0.2\%\\
German & \cellcolor{cyan!0} 0.0\% & \cellcolor{cyan!0} 0.0\% & \cellcolor{cyan!0} 0.0\% & \cellcolor{cyan!27} 1.1\% & \cellcolor{cyan!0} 0.0\% & \cellcolor{cyan!2} 0.1\% &  & \cellcolor{cyan!0} 0.0\% & \cellcolor{cyan!0} 0.0\% & \cellcolor{cyan!0} 0.0\% & \cellcolor{cyan!20} 0.8\% & \cellcolor{cyan!0} 0.0\% & \cellcolor{cyan!5} 0.2\%\\
Farsi & \cellcolor{cyan!0} 0.0\% & \cellcolor{cyan!0} 0.0\% & \cellcolor{cyan!0} 0.0\% & \cellcolor{cyan!30} 1.2\% & \cellcolor{cyan!0} 0.0\% & \cellcolor{cyan!0} 0.0\% &  & \cellcolor{cyan!0} 0.0\% & \cellcolor{cyan!0} 0.0\% & \cellcolor{cyan!0} 0.0\% & \cellcolor{cyan!27} 1.1\% & \cellcolor{cyan!0} 0.0\% & \cellcolor{cyan!2} 0.1\%\\
Telugu & \cellcolor{cyan!0} 0.0\% & \cellcolor{cyan!0} 0.0\% & \cellcolor{cyan!0} 0.0\% & \cellcolor{cyan!43} 1.7\% & \cellcolor{cyan!0} 0.0\% & \cellcolor{cyan!2} 0.1\% &  & \cellcolor{cyan!0} 0.0\% & \cellcolor{cyan!5} 0.2\% & \cellcolor{cyan!0} 0.0\% & \cellcolor{cyan!45} 1.8\% & \cellcolor{cyan!0} 0.0\% & \cellcolor{cyan!2} 0.1\%\\
Ukranian & \cellcolor{cyan!0} 0.0\% & \cellcolor{cyan!0} 0.0\% & \cellcolor{cyan!0} 0.0\% & \cellcolor{cyan!22} 0.9\% & \cellcolor{cyan!0} 0.0\% & \cellcolor{cyan!0} 0.0\% &  & \cellcolor{cyan!0} 0.0\% & \cellcolor{cyan!0} 0.0\% & \cellcolor{cyan!0} 0.0\% & \cellcolor{cyan!32} 1.3\% & \cellcolor{cyan!0} 0.0\% & \cellcolor{cyan!0} 0.0\%\\
Czech & \cellcolor{cyan!0} 0.0\% & \cellcolor{cyan!0} 0.0\% & \cellcolor{cyan!0} 0.0\% & \cellcolor{cyan!27} 1.1\% & \cellcolor{cyan!0} 0.0\% & \cellcolor{cyan!0} 0.0\% &  & \cellcolor{cyan!0} 0.0\% & \cellcolor{cyan!0} 0.0\% & \cellcolor{cyan!0} 0.0\% & \cellcolor{cyan!75} 3.0\% & \cellcolor{cyan!0} 0.0\% & \cellcolor{cyan!0} 0.0\%\\
\bottomrule
\end{tabular}

\vspace{-2mm}
\caption{Response error rate for the \textbf{feedback selection} task.}
\medskip
\begin{tabular}{ lrrrrrr@{\hspace{2mm}}crrrrrr }
\toprule
& \multicolumn{ 6 }{c}{\bf English prompt} && \multicolumn{ 6 }{c}{\bf Translated prompt} \\
\textbf{Language} & \bf GPT4o & \bf LLama & \bf Claude & \bf Gemini & \bf Mistral & \bf Cmd-A & \bf  & \bf GPT4o & \bf LLama & \bf Claude & \bf Gemini & \bf Mistral & \bf Cmd-A\\
\midrule
English & \cellcolor{cyan!3} 23.7\% & \cellcolor{cyan!18} 45.8\% & \cellcolor{cyan!39} 75.0\% & \cellcolor{cyan!6} 27.8\% & \cellcolor{cyan!2} 23.0\% & \cellcolor{cyan!11} 35.1\% &  & \cellcolor{cyan!3} 23.7\% & \cellcolor{cyan!18} 45.8\% & \cellcolor{cyan!39} 75.0\% & \cellcolor{cyan!6} 27.8\% & \cellcolor{cyan!2} 23.0\% & \cellcolor{cyan!11} 35.1\%\\
Mandarin & \cellcolor{cyan!5} 26.9\% & \cellcolor{cyan!25} 54.5\% & \cellcolor{cyan!44} 81.5\% & \cellcolor{cyan!12} 36.9\% & \cellcolor{cyan!10} 33.8\% & \cellcolor{cyan!20} 47.7\% &  & \cellcolor{cyan!9} 32.6\% & \cellcolor{cyan!37} 71.9\% & \cellcolor{cyan!50} 89.7\% & \cellcolor{cyan!3} 24.3\% & \cellcolor{cyan!21} 49.3\% & \cellcolor{cyan!25} 54.5\%\\
Hindi & \cellcolor{cyan!7} 30.3\% & \cellcolor{cyan!16} 42.3\% & \cellcolor{cyan!43} 79.6\% & \cellcolor{cyan!11} 34.9\% & \cellcolor{cyan!7} 29.9\% & \cellcolor{cyan!20} 47.9\% &  & \cellcolor{cyan!25} 55.5\% & \cellcolor{cyan!42} 78.9\% & \cellcolor{cyan!48} 87.7\% & \cellcolor{cyan!9} 33.2\% & \cellcolor{cyan!35} 68.9\% & \cellcolor{cyan!37} 71.3\%\\
Arabic & \cellcolor{cyan!6} 28.0\% & \cellcolor{cyan!24} 54.1\% & \cellcolor{cyan!42} 79.5\% & \cellcolor{cyan!11} 35.3\% & \cellcolor{cyan!9} 32.9\% & \cellcolor{cyan!17} 44.0\% &  & \cellcolor{cyan!1} 21.9\% & \cellcolor{cyan!44} 81.7\% & \cellcolor{cyan!38} 73.6\% & \cellcolor{cyan!2} 22.4\% & \cellcolor{cyan!12} 36.4\% & \cellcolor{cyan!21} 49.5\%\\
German & \cellcolor{cyan!4} 25.1\% & \cellcolor{cyan!21} 48.8\% & \cellcolor{cyan!42} 79.4\% & \cellcolor{cyan!9} 32.7\% & \cellcolor{cyan!7} 30.4\% & \cellcolor{cyan!18} 45.4\% &  & \cellcolor{cyan!2} 22.3\% & \cellcolor{cyan!25} 54.5\% & \cellcolor{cyan!41} 77.0\% & \cellcolor{cyan!7} 29.6\% & \cellcolor{cyan!9} 32.9\% & \cellcolor{cyan!11} 36.0\%\\
Farsi & \cellcolor{cyan!6} 28.5\% & \cellcolor{cyan!23} 52.5\% & \cellcolor{cyan!45} 82.5\% & \cellcolor{cyan!8} 31.6\% & \cellcolor{cyan!9} 32.6\% & \cellcolor{cyan!18} 45.5\% &  & \cellcolor{cyan!1} 21.6\% & \cellcolor{cyan!23} 52.1\% & \cellcolor{cyan!39} 75.1\% & \cellcolor{cyan!7} 29.3\% & \cellcolor{cyan!7} 30.3\% & \cellcolor{cyan!6} 28.9\%\\
Telugu & \cellcolor{cyan!7} 29.2\% & \cellcolor{cyan!25} 55.6\% & \cellcolor{cyan!44} 81.5\% & \cellcolor{cyan!11} 35.2\% & \cellcolor{cyan!9} 33.1\% & \cellcolor{cyan!23} 52.4\% &  & \cellcolor{cyan!42} 78.3\% & \cellcolor{cyan!38} 73.8\% & \cellcolor{cyan!50} 89.5\% & \cellcolor{cyan!13} 37.9\% & \cellcolor{cyan!36} 70.9\% & \cellcolor{cyan!42} 78.8\%\\
Ukranian & \cellcolor{cyan!5} 27.3\% & \cellcolor{cyan!21} 49.4\% & \cellcolor{cyan!43} 80.3\% & \cellcolor{cyan!9} 32.7\% & \cellcolor{cyan!10} 33.5\% & \cellcolor{cyan!18} 45.2\% &  & \cellcolor{cyan!19} 47.3\% & \cellcolor{cyan!36} 69.7\% & \cellcolor{cyan!48} 87.1\% & \cellcolor{cyan!0} 20.7\% & \cellcolor{cyan!19} 46.7\% & \cellcolor{cyan!21} 49.7\%\\
Czech & \cellcolor{cyan!6} 27.9\% & \cellcolor{cyan!14} 39.5\% & \cellcolor{cyan!43} 80.2\% & \cellcolor{cyan!7} 30.2\% & \cellcolor{cyan!8} 31.8\% & \cellcolor{cyan!21} 49.0\% &  & \cellcolor{cyan!11} 34.9\% & \cellcolor{cyan!28} 59.5\% & \cellcolor{cyan!34} 67.8\% & \cellcolor{cyan!2} 23.0\% & \cellcolor{cyan!9} 33.1\% & \cellcolor{cyan!13} 38.0\%\\
\bottomrule
\end{tabular}

\vspace{-2mm}
\caption{Rate of defaulting to the correct answer for the 
\textbf{feedback selection} task.}
\label{tab:default_correct}
\medskip
\begin{tabular}{ lrrrrrr@{\hspace{2mm}}crrrrrr }
\toprule
& \multicolumn{ 6 }{c}{\bf English prompt} && \multicolumn{ 6 }{c}{\bf Translated prompt} \\
\textbf{Language} & \bf GPT4o & \bf LLama & \bf Claude & \bf Gemini & \bf Mistral & \bf Cmd-A & \bf  & \bf GPT4o & \bf LLama & \bf Claude & \bf Gemini & \bf Mistral & \bf Cmd-A\\
\midrule
Mandarin & \cellcolor{green!0} 0.0\% & \cellcolor{green!2} 0.1\% & \cellcolor{green!12} 0.5\% & \cellcolor{green!0} 0.0\% & \cellcolor{green!0} 0.0\% & \cellcolor{green!0} 0.0\% &  & \cellcolor{green!0} 0.0\% & \cellcolor{green!2} 0.1\% & \cellcolor{green!100} 4.2\% & \cellcolor{green!0} 0.0\% & \cellcolor{green!0} 0.0\% & \cellcolor{green!0} 0.0\%\\
Hindi & \cellcolor{green!0} 0.0\% & \cellcolor{green!0} 0.0\% & \cellcolor{green!0} 0.0\% & \cellcolor{green!0} 0.0\% & \cellcolor{green!0} 0.0\% & \cellcolor{green!0} 0.0\% &  & \cellcolor{green!0} 0.0\% & \cellcolor{green!0} 0.0\% & \cellcolor{green!5} 0.2\% & \cellcolor{green!0} 0.0\% & \cellcolor{green!0} 0.0\% & \cellcolor{green!0} 0.0\%\\
Arabic & \cellcolor{green!0} 0.0\% & \cellcolor{green!0} 0.0\% & \cellcolor{green!0} 0.0\% & \cellcolor{green!0} 0.0\% & \cellcolor{green!0} 0.0\% & \cellcolor{green!0} 0.0\% &  & \cellcolor{green!0} 0.0\% & \cellcolor{green!0} 0.0\% & \cellcolor{green!92} 3.7\% & \cellcolor{green!0} 0.0\% & \cellcolor{green!0} 0.0\% & \cellcolor{green!0} 0.0\%\\
German & \cellcolor{green!0} 0.0\% & \cellcolor{green!0} 0.0\% & \cellcolor{green!2} 0.1\% & \cellcolor{green!0} 0.0\% & \cellcolor{green!0} 0.0\% & \cellcolor{green!0} 0.0\% &  & \cellcolor{green!0} 0.0\% & \cellcolor{green!0} 0.0\% & \cellcolor{green!88} 3.5\% & \cellcolor{green!0} 0.0\% & \cellcolor{green!0} 0.0\% & \cellcolor{green!0} 0.0\%\\
Farsi & \cellcolor{green!0} 0.0\% & \cellcolor{green!0} 0.0\% & \cellcolor{green!0} 0.0\% & \cellcolor{green!0} 0.0\% & \cellcolor{green!0} 0.0\% & \cellcolor{green!0} 0.0\% &  & \cellcolor{green!0} 0.0\% & \cellcolor{green!0} 0.0\% & \cellcolor{green!0} 0.0\% & \cellcolor{green!0} 0.0\% & \cellcolor{green!0} 0.0\% & \cellcolor{green!0} 0.0\%\\
Telugu & \cellcolor{green!0} 0.0\% & \cellcolor{green!0} 0.0\% & \cellcolor{green!0} 0.0\% & \cellcolor{green!0} 0.0\% & \cellcolor{green!0} 0.0\% & \cellcolor{green!0} 0.0\% &  & \cellcolor{green!0} 0.0\% & \cellcolor{green!0} 0.0\% & \cellcolor{green!12} 0.5\% & \cellcolor{green!0} 0.0\% & \cellcolor{green!0} 0.0\% & \cellcolor{green!0} 0.0\%\\
Ukranian & \cellcolor{green!0} 0.0\% & \cellcolor{green!0} 0.0\% & \cellcolor{green!0} 0.0\% & \cellcolor{green!0} 0.0\% & \cellcolor{green!0} 0.0\% & \cellcolor{green!0} 0.0\% &  & \cellcolor{green!0} 0.0\% & \cellcolor{green!0} 0.0\% & \cellcolor{green!38} 1.5\% & \cellcolor{green!0} 0.0\% & \cellcolor{green!0} 0.0\% & \cellcolor{green!0} 0.0\%\\
Czech & \cellcolor{green!0} 0.0\% & \cellcolor{green!0} 0.0\% & \cellcolor{green!0} 0.0\% & \cellcolor{green!0} 0.0\% & \cellcolor{green!0} 0.0\% & \cellcolor{green!0} 0.0\% &  & \cellcolor{green!0} 0.0\% & \cellcolor{green!0} 0.0\% & \cellcolor{green!70} 2.8\% & \cellcolor{green!0} 0.0\% & \cellcolor{green!0} 0.0\% & \cellcolor{green!0} 0.0\%\\
\bottomrule
\end{tabular}
\vspace{-2mm}
\caption{Response error rate for the \textbf{translation grading} task.}

\vspace{-15mm}
\end{table*}

\clearpage

\section{Experiment Prompts}
\subsection{Task: Misconception Identification}
We used a sequence of 3 prompts:\\
\begin{lstlisting}[style=promptstyle,mathescape=true]
$\textbf{System prompt:}$
You are an expert math tutor who knows about all grade-school level math misconceptions. Your task is to select the accurate type of misconceptions your student has based on the (incorrect) answer he/she gives to a multiple-choice math question. You will be given 4 misconceptions types. Your selected misconception type should correspond to the given question and answer. Explain your reasoning
\end{lstlisting}

\begin{lstlisting}[style=promptstyle,mathescape=true]
$\textbf{User message 1:}$
Question: {QUESTION}
Selected Answer: {SELECTED_ANSWER}
Misconceptions:
A. {Misconception 1}
B. {Misconception 2}
C. {Misconception 3}
D. {Misconception 4}
\end{lstlisting}

\noindent The position of the Misconception corresponding to the selected answer rotates from question to question. The subsequent assistant message is stored as the chain-of-thought. Thereafter, we sent the second user message.

\begin{lstlisting}[style=promptstyle,mathescape=true]
$\textbf{User message 2:}$
Now based on your above explanation, output the option corresponding to the correct misconception. Only say 'A', 'B', 'C', or 'D' without any other text. Do not say anything else.
\end{lstlisting}
The response to this part is the final answer. We regenerate until an answer of `A', `B', `C', or `D' is received, up to 20 times. If no answer is received, a response of `E' is saved.

This method is used for all models except Gemini. In case of Gemini, we use the \texttt{generate\_content} method, which is recommended for non-chat tasks and allows for a single user message. In this case, after obtaining the chain-of-thought, we make a new query with the same system prompt but with the following user message:\\

\begin{lstlisting}[style=promptstyle,mathescape=true]
$\textbf{Gemini message:}$
You have previously given the following answer and explanation:
{COT}
Now based on your above explanation, output the option corresponding to the correct misconception. Only say 'A', 'B', 'C', or 'D' without any other text. Do not say anything else.
\end{lstlisting}
\noindent Note that the last part is identical to User Message 2

\noindent When using translated prompts, the System Prompt and, User Message 2 and Gemini Message are translated to the target language.

\clearpage

\subsection{Task: Feedback Selection}
\begin{lstlisting}[style=promptstyle,mathescape=true]
$\textbf{System prompt:}$
You are an expert math tutor who specialises in providing precise and helpful feedback for grade-school level math questions. Your task is to select the correct explanation for a student's given answer to a multiple-choice math question.

You will be provided with:
- A math question
- A specific answer chosen by the student (which can be correct or incorrect).
- Four possible explanations (labelled A, B, C, and D). 
Your selected explanation should accurately correspond to the given answer. Provide your reasoning for selecting the explanation.
\end{lstlisting}
\begin{lstlisting}[style=promptstyle,mathescape=true]
$\textbf{User message 1:}$
Question: {QUESTION}
Selected Answer: {SELECTED_ANSWER}
Feedbacks:
A. {Feedback 1}
B. {Feedback 2}
C. {Feedback 3}
D. {Feedback 4}
\end{lstlisting}

\noindent The position of the Feedback corresponding to the selected answer rotates from question to question. If it is placed at positions A, B, or C, the feedback corresponding to the correct answer is at position D. Otherwise, it is at C. The subsequent assistant message is stored as the chain-of-thought. Thereafter, we sent the second user message.

\begin{lstlisting}[style=promptstyle,mathescape=true]
$\textbf{User message 2:}$
Now based on your above explanation, output the option corresponding to the correct explanation. Only say 'A', 'B', 'C', or 'D' without any other text. Do not say anything else.
\end{lstlisting}
The response to this part is the final answer. We regenerate until an answer of `A', `B', `C', or `D' is received, up to 20 times. If no answer is received, a response of `E' is saved.

This method is used for all models except Gemini. In case of Gemini, we use the \texttt{generate\_content} method, which is recommended for non-chat tasks and allows for a single user message. In this case, after obtaining the chain-of-thought, we make a new query with the same system prompt but with the following user message:\\
\begin{lstlisting}[style=promptstyle,mathescape=true]
$\textbf{Gemini message:}$
You have previously given the following answer and explanation:
{COT}
Now based on your above explanation, output the option corresponding to the correct explanation. Only say 'A', 'B', 'C', or 'D' without any other text. Do not say anything else.
\end{lstlisting}
\noindent Note that the last part is identical to User Message 2.
When using translated prompts, the System Prompt, User Message 2, and Gemini Message are translated to the target language. We manually made sure that the formatting was maintained after the translation.

\clearpage

\subsection{Task: Tutoring}
\begin{lstlisting}[style=promptstyle,mathescape=true]
$\textbf{Student system prompt:}$
Student Persona: {STUDENT_PERSONA}

Math problem: {MATH_PROBLEM}

Student solution: {STUDENT_SOLUTION}

Context: You need to role-play the student, {STUDENT_NAME}, while the user roleplays the tutor. {STUDENT_NAME} thinks their answer is correct. Only when the teacher provides several good reasoning questions, {STUDENT_NAME} understands the problem and corrects the solution. {STUDENT_NAME} can use calculator and thus makes no calculation errors. Send <EOM> tag at end of the student message.
\end{lstlisting}

\begin{lstlisting}[style=promptstyle,mathescape=true]
$\textbf{Teacher system prompt:}$
A tutor and a student work together to solve the following math word problem. 
Math problem: {MATH_PROBLEM}
The correct solution is as follows:
{GROUND_TRUTH}

You need to role-play the tutor while the user roleplays the student, {STUDENT_NAME}. The tutor is a soft-spoken empathetic man who dislikes giving out direct answers to students, and instead likes to answer questions with other questions that would help the student understand the concepts, so that she can solve the problem themselves. 

{STUDENT_NAME} has come up with a solution, but it is incorrect. Please start the conversation, one line at a time, aiming to figure out what is {STUDENT_NAME}'s solution and what is wrong with it. Then try to get her to fix it.
\end{lstlisting}

The dialogue history was formatted as user-assistant message pairs for teacher and student roles. We manually set the initial messages to initiate conversations in the target language.
\clearpage

\subsection{Task: Translation Grading}
\begin{lstlisting}[style=promptstyle,mathescape=true]
$\textbf{System prompt:}$
You are a language translation evaluator. Your task is to assess the quality of a translation from English to {LANGUAGE}. You will be provided with two sentences:
1. An original English sentence.
2. A translated sentence in {LANGUAGE}.
Your goal is to rate the translation on a scale from 1 to 5 based on the following criteria:
1: The translation is incorrect, incomprehensible, or completely unrelated to the original English sentence.
2: The translation has significant errors and distorts the meaning of the original English sentence.
3: The translation is understandable but contains notable errors or awkward phrasing.
4: The translation is mostly accurate with minor errors or slightly awkward phrasing.
5: The translation is fluent, natural, and accurately conveys the meaning of the original English sentence without errors.
Explain your decision
\end{lstlisting}
\begin{lstlisting}[style=promptstyle,mathescape=true]
$\textbf{User message 1:}$
English: {ENGLISH_SENTENCE}
{LANGUAGE}: {TRANSLATED_SENTENCE}
\end{lstlisting}
\noindent  The subsequent assistant message is stored as the chain-of-thought. Thereafter, we sent the second user message.

\begin{lstlisting}[style=promptstyle,mathescape=true]
$\textbf{User message 2:}$
Now based on your above explanation, output the final score from 1 to 5. Only say '1', '2', '3', '4', or '5' without any other text. Do not say anything else.
\end{lstlisting}
\noindent The response to this part is the final answer. We regenerate until an answer of `1', `2', `3', `4', or `5' is received, up to 20 times. If no answer is received, a response of `0' is saved.

This method is used for all models except Gemini. In case of Gemini, we use the \texttt{generate\_content} method, which is recommended for non-chat tasks and allows for a single user message. In this case, after obtaining the chain-of-thought, we make a new query with the same system prompt but with the following user message:\\
\begin{lstlisting}[style=promptstyle,mathescape=true]
$\textbf{Gemini message:}$
You have previously given the following answer and explanation:
{COT}
Now based on your above explanation, output the final score from 1 to 5. Only say '1', '2', '3', '4', or '5' without any other text. Do not say anything else.
\end{lstlisting}
\noindent Note that the last part is identical to User Message 2

This sequence is repeated twice for each sentence, once with the original translation and once with the perturbed translation. The scores are then compared. When using English prompts, the LANGUAGE fields are set to their English exonyms, i.e., Mandarin, Hindi, Arabic, German, Farsi, Telugu, Ukrainian, and Czech. When using translated prompts, the System Prompt, User Message 2, and Gemini Message are translated to the target language. We manually made sure that the formatting was maintained after the translation. We also use the language endonyms, namely \raisebox{-1.4mm}{\includegraphics[height=1em]{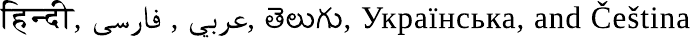}}.

\clearpage

\section{Translation Quality}
\label{sec:traqual}
As we mentioned in Limitations, an LLM performing poorly in a given language does not necessarily mean that the LLM itself is bad.
It could also mean that information was lost during translation.
This is particularly problematic because the machine translation systems likely suffer from the same resource limitations that plague the LLMs in the first place.
As such, we manually investigated a small subset of translated questions for the languages they we are fluent in, namely Persian, Arabic, Czech, and Hindi.
For each language, we analysed 10 questions each for the Feedback and Misconception tasks, and 20 questions for the Translation Grading task.

In the case of Persian, the only recurring error was with mathematical notation, particularly that the minus sign gets placed to the right of the numbers instead of the left, where it should be. This, however, seems to be a rendering issue, which is a result of the fact that the minus sign (`$-$', U+2212) is often replaced by the similar-looking hyphen (`-', U+002D), confusing the rendering program into believing that it is rendering text.
This should not be an issue since LLMs take raw Unicode encodings as input.
Beyond this, there were some minor tense errors, but the meanings were clear.

The issue with sign placement was also observed in Arabic. In addition, there seem to be some translation errors. For example, the word `travel' used here in the context of the movement of a graph was translated to `liyusaafir', which is more like `taking a trip'. We found no errors in the sentences for the translation task.
In Czech, the primary source of errors was improper context-dependent terminology.
For example, when translating the word `co-interior (angles)', it missed the `co' prefix and translated only the `interior' part.
While this is fine in regular speech, in Mathematical terminology, this can be confusing.
Despite making the translation harder to follow, the core meaning of the question is preserved.

In Hindi we found several cases where the Hindi sentence was difficult to follow for the Hindi speaking author due to misinterpretation of polysemes by the translator e.g. the word `round', which was being used in the sense of `approximate' was translated to the sense of `circle' and `property' which was being used in the sense of `quality', was translated as `possessions'. Also, the phrase `Not Quite' was translated to something like `Not Enough', perhaps due to the word `quite' not having a Hindi equivalent. However, given the context, using the word for 'Almost' would have been more tonally accurate. However, quite a few translations were hard for the annotator to follow, but backtranslating them yielded reasonably good results, meaning there was no information loss. 

The translation exercises showed few errors, perhaps due to the sentences being easy to translate by design. There were one or two mistranslations, but otherwise it worked well. One minor issue was that word boundary detection, which was performed in Python using the regex \texttt{`\textbackslash b\textbackslash w+\textbackslash b'}, sometimes identified individual characters in Hindi rather than whole words. However, the resulting sentence still had errors, just not the type of errors that we expected.


\begin{figure*}[ht]
\centering
\bf Results for prompt in English \\

\begin{minipage}{0.14\linewidth}
\bf Task 1:\\
\bf Misconception identification
\vspace{50mm}
\end{minipage}
\includegraphics[width=0.85\linewidth,trim={7mm 0 10mm 0},clip]{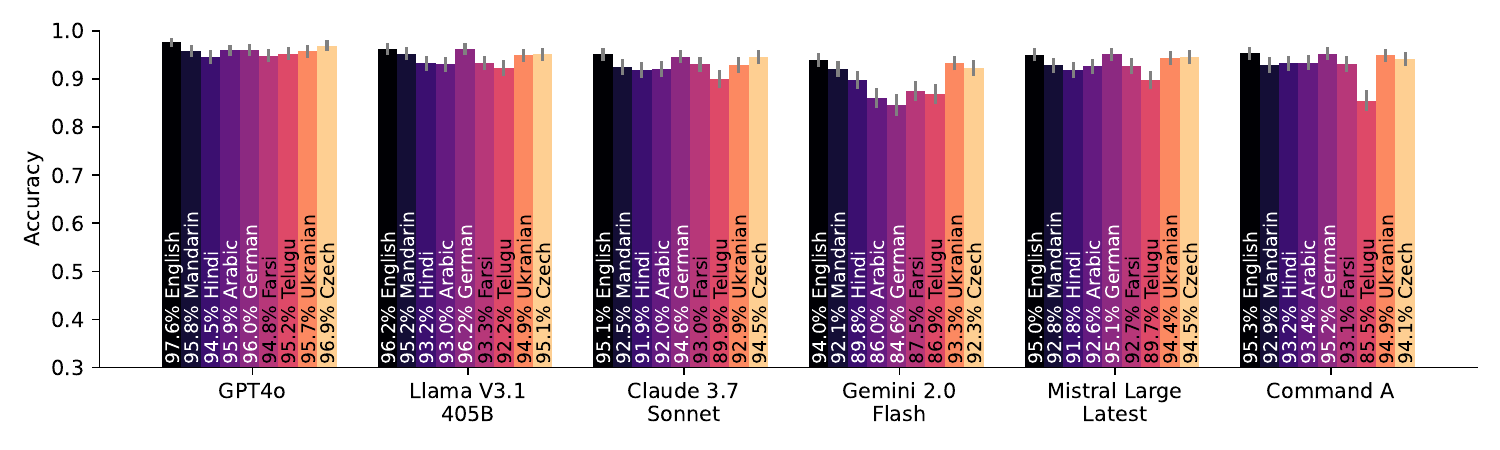}
\vspace{-33mm}

\begin{minipage}{0.14\linewidth}
\bf Task 2:\\
\bf Feedback \,\,\, selection
\vspace{50mm}
\end{minipage}
\includegraphics[width=0.85\linewidth,trim={7mm 0 10mm 0},clip]{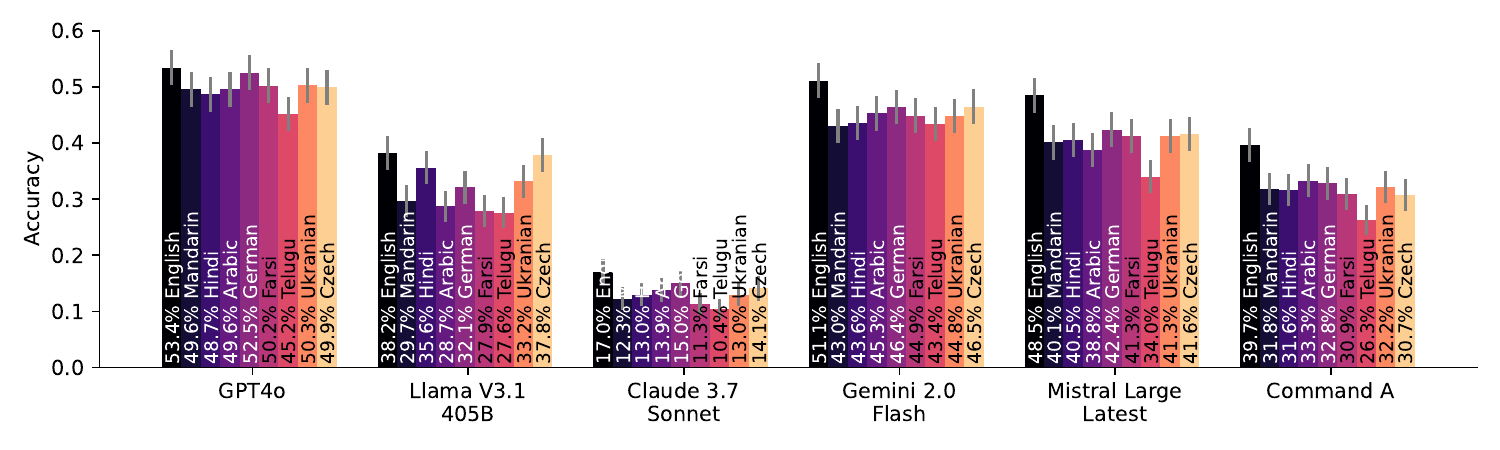}
\vspace{-33mm}

\begin{minipage}{0.14\linewidth}
\bf Task 3:\\
\bf Tutoring
\vspace{50mm}
\end{minipage}
\includegraphics[width=0.85\linewidth,trim={7mm 0 10mm 0},clip]{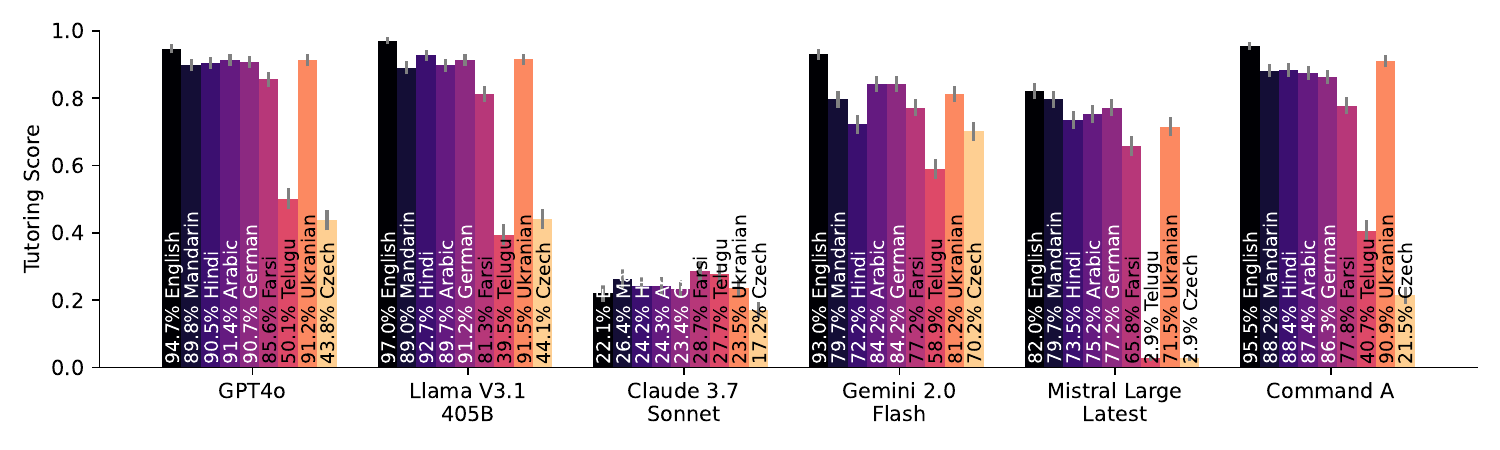}
\vspace{-30mm}

\begin{minipage}{0.14\linewidth}
\bf Task 4:\\
\bf Translation grading
\vspace{50mm}
\end{minipage}
\includegraphics[width=0.85\linewidth,trim={7mm 0 10mm 0},clip]{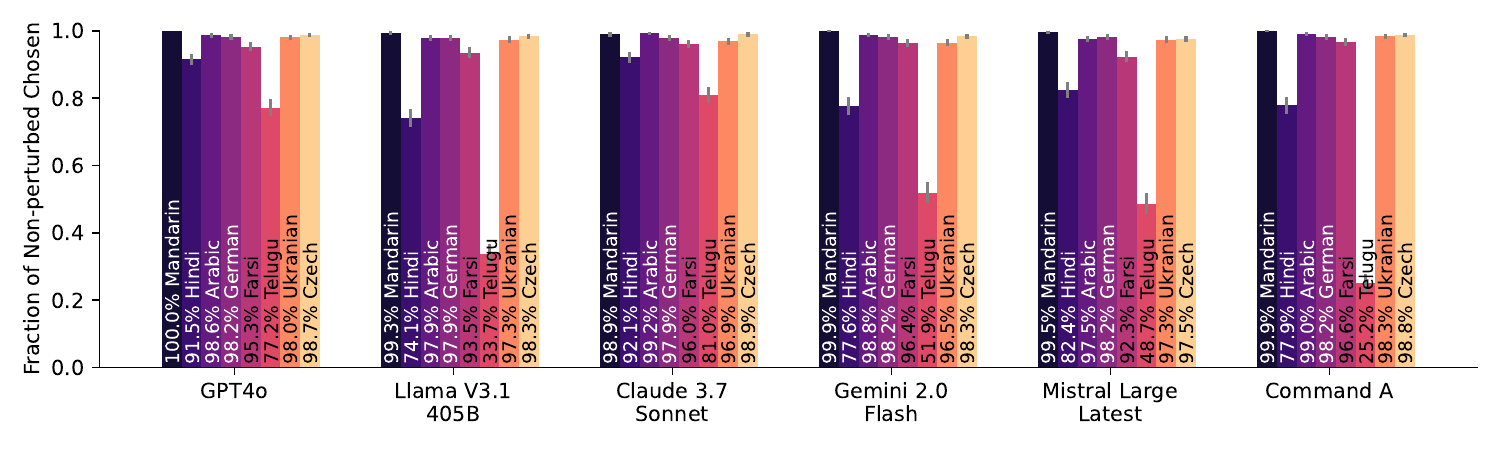}
\vspace{-30mm}

\caption{
Evaluation results of the four tasks across five lare language models.
The error bars show a 95\% confidence interval (t-test).
MathDial Graphs show \textit{tutoring score after five turns}, most models flatline after 5 utterance pairs.
The English language column is absent because translation evaluation uses English as the source.
All scores range from 0.0 to 1.0, with higher being better, though they are not comparable with each other.
Note the truncated y-axes for better detail.
Visualizes \Cref{tab:result_misconception,tab:result_feedback,tab:result_translation,tab:result_tutoring}.
}
\label{fig:results_english}
\end{figure*}

\begin{figure*}[ht]
\centering
\bf Results for prompt in target language \\

\begin{minipage}{0.14\linewidth}
\bf Task 1:\\
\bf Misconception identification
\vspace{50mm}
\end{minipage}
\includegraphics[width=0.85\linewidth,trim={7mm 0 10mm 0},clip]{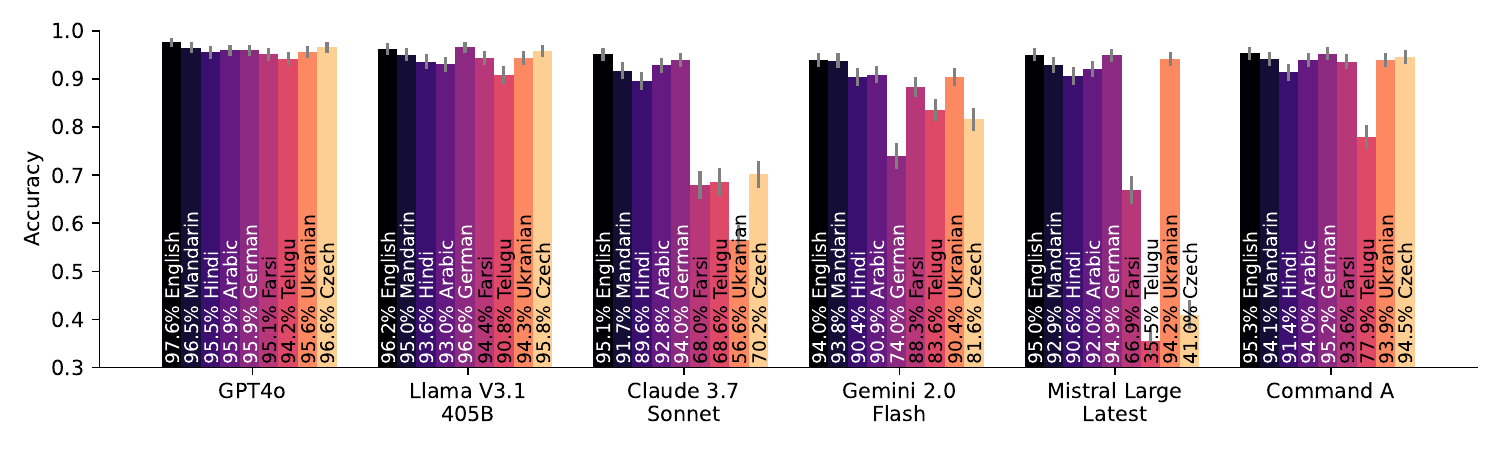}
\vspace{-33mm}

\begin{minipage}{0.14\linewidth}
\bf Task 2:\\
\bf Feedback \,\,\, selection
\vspace{50mm}
\end{minipage}
\includegraphics[width=0.85\linewidth,trim={7mm 0 10mm 0},clip]{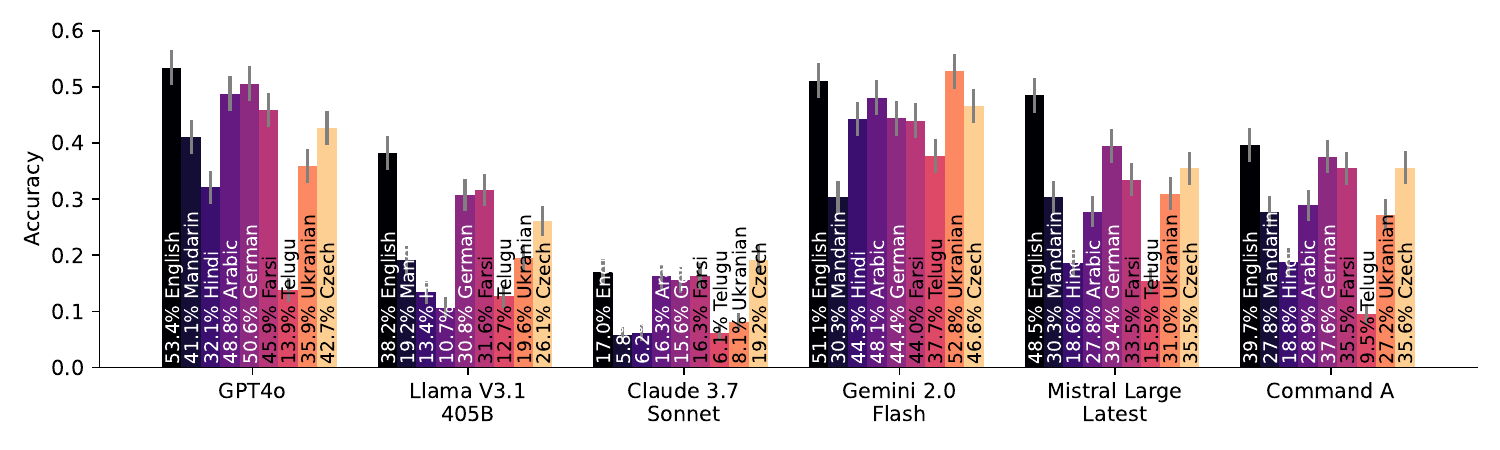}
\vspace{-33mm}

\begin{minipage}{0.14\linewidth}
\bf Task 4:\\
\bf Translation grading
\vspace{50mm}
\end{minipage}
\includegraphics[width=0.85\linewidth,trim={7mm 0 10mm 0},clip]{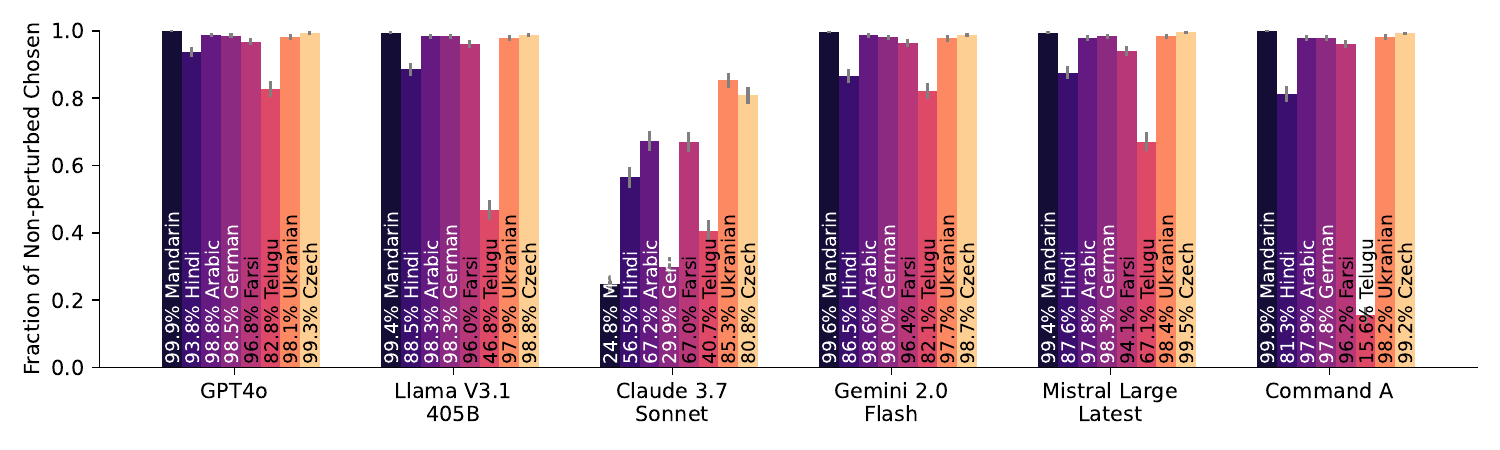}
\vspace{-30mm}

\caption{
Evaluation results of the four tasks across five large language models.
The error bars show 95\% confidence interval (t-test).
MathDial Graphs show \textit{tutoring score after five turns}, most models flatline after 5 utterance pairs.
The English language column is absent because translation evaluation uses English as the source.
All scores range from 0.0 to 1.0, with higher being better, though they are not comparable with each other.
Note the truncated y-axes for better detail.
Visualizes \Cref{tab:result_misconception,tab:result_feedback,tab:result_translation,tab:result_tutoring}.
}
\label{fig:results_native}
\end{figure*}

\end{document}